\documentclass[10pt,twocolumn,letterpaper]{article}

\usepackage{iccv}

\makeatletter
\@namedef{ver@everyshi.sty}{}
\makeatother
\usepackage{tikz}

\usepackage{times}
\usepackage{epsfig}
\usepackage{graphicx}
\usepackage{amsmath}
\usepackage{amssymb}
\usepackage{todonotes}
\usepackage{multirow}
\usepackage{booktabs}
\usepackage{multirow}
\usepackage{multicol}
\usepackage{comment}
\usepackage{bm}
\usepackage{caption}
\usepackage{cite}
\usepackage{algorithmic}
\usepackage{algorithm}
\usepackage{color}
\usepackage{colortbl}
\usepackage{cite}
\usepackage{balance}
\usepackage{siunitx}
\usepackage{flafter}
\usepackage{makecell}
\usepackage{array}

\usepackage[pagebackref=true,breaklinks=true,letterpaper=true,colorlinks,bookmarks=false]{hyperref}

\usepackage[capitalize]{cleveref}
\crefname{section}{Sec.}{Secs.}
\Crefname{section}{Section}{Sections}
\Crefname{table}{Table}{Tables}
\crefname{table}{Tab.}{Tabs.}

\def\vct#1{\mbox{\boldmath $#1$}}
\def\fd{f_\text{d}}
\def\fmlp{f_\text{mlp}}
\DeclareMathOperator*{\argmin}{arg\,min}

\iccvfinalcopy % *** Uncomment this line for the final submission

\def\iccvPaperID{4654} % *** Enter the ICCV Paper ID here

\ificcvfinal\fi

\begin{document}

%%%%%%%%% TITLE
%\title{TransPoser: A Deep Generative Optimization Model for\\Joint Object Pose and Shape Estimations from Online Multi-view Observations}
\title{TransPoser: Transformer as an Optimizer \\for Joint Object Shape and Pose Estimation}

\author{Yuta Yoshitake \;\;\;\;\;\; Mai Nishimura \;\;\;\;\;\; Shohei Nobuhara\;\;\;\;\;\;Ko Nishino\\
Graduate School of Informatics, Kyoto University, Kyoto, Japan\\
%{\tt\small \{nob,kon\}@i.kyoto-u.ac.jp}
}

%\and

%\maketitle
\ificcvfinal\thispagestyle{empty}\fi

\twocolumn[{
\maketitle
\begin{center}
    \vspace{-7mm}
    \captionsetup{type=figure}
        \includegraphics[width=\linewidth]{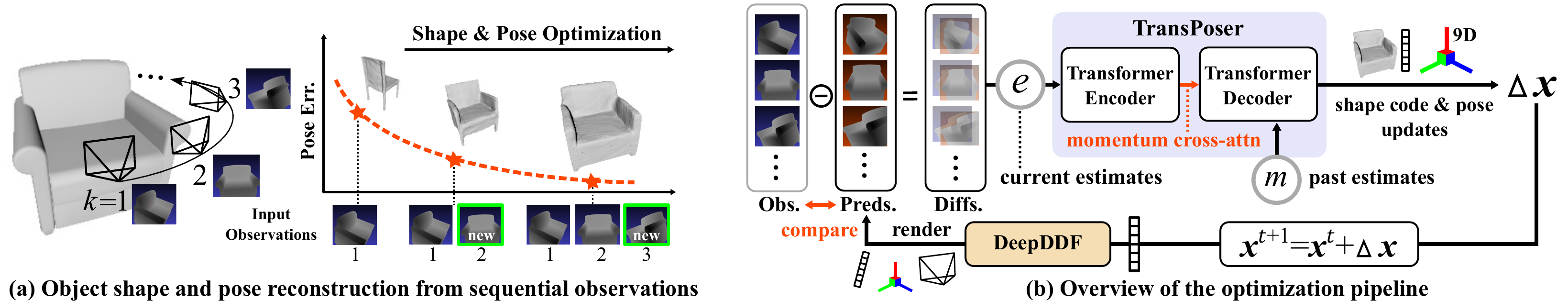}
    \captionof{figure}{We propose TransPoser, a novel neural optimization network for joint object pose and shape reconstruction from sequential observations (a). We also derive DeepDDF, a novel category-level neural 3D shape representation that directly outputs 2D depth images for given viewing conditions. By integrating DeepDDF and TransPoser, we realize efficient and accurate learned weighted optimization with learned momentum for joint shape and pose estimation (b).}
    \label{fig:opening}
\end{center}
}]

%%%%%%%%% ABSTRACT
\begin{abstract}

We propose a novel method for joint estimation of shape and pose of rigid objects from their sequentially observed RGB-D images. In sharp contrast to past approaches that rely on complex non-linear optimization, we propose to formulate it as a neural optimization that learns to efficiently estimate the shape and pose. We introduce Deep Directional Distance Function (DeepDDF), a neural network that directly outputs the depth image of an object given the camera viewpoint and viewing direction, for efficient error computation in 2D image space. We formulate the joint estimation itself as a Transformer which we refer to as TransPoser. We fully leverage the tokenization and multi-head attention to sequentially process the growing set of observations and to efficiently update the shape and pose with a learned momentum, respectively. Experimental results on synthetic and real data show that DeepDDF achieves high accuracy as a category-level object shape representation and TransPoser achieves state-of-the-art accuracy efficiently for joint shape and pose estimation.

\end{abstract}

%%%%%%%%% BODY TEXT
\begin{table*}[t!]
    \centering
    \begingroup
    \renewcommand{\arraystretch}{0.85}
    \small
    \begin{tabular}{l|ccccccc}
        \toprule[1.5pt]
        \multirow{2}{*}{Method} &
        \multirow{2}{*}{\!\begin{tabular}{c}Feedback\\Loop\end{tabular}\!} &
        \multirow{2}{*}{\!\begin{tabular}{c}Fewer\\Updates\end{tabular}\!} &
        \multirow{2}{*}{\begin{tabular}{c}Joint Shape and\\Pose Estimation\end{tabular}} &
        \multirow{2}{*}{\begin{tabular}{c}Multi-view\\Observation\end{tabular}} &
        \multirow{2}{*}{Online} &
        \multirow{2}{*}{\begin{tabular}{c}Shape\\Representation\end{tabular}} &
        \multirow{2}{*}{Momentum} \\ \\
        \hline
         &  &  &  &  &  &  & \vspace{-7pt} \\
        DPDN \cite{lin2022category} &  & - & \checkmark &  & - & \footnotesize{Point Cloud} & - \\
        Scan2CAD \cite{Avetisyan_2019_CVPR} &  & - & \checkmark & \checkmark &   & \footnotesize{CAD models} & - \\
        RayTran \cite{tyszkiewicz2022raytran} &  & - & \checkmark & \checkmark &   & \footnotesize{Voxel Occupancy} & - \\
        Vid2CAD \cite{maninis2022vid2cad} & \checkmark &  & \checkmark & \checkmark & \checkmark & \footnotesize{CAD models} & \footnotesize{Moving Avg.} \\
        ELLIPSDF \cite{shan2021ellipsdf} & \checkmark &  & \checkmark & \checkmark & \checkmark & \footnotesize{DeepSDF} & \footnotesize{Moving Avg.} \\
        Ma \etal \cite{ma2020deep} & \checkmark & \checkmark &  &  &  & - &  \\
        \hline
         &  &  &  &  &  &  & \vspace{-7pt} \\
        \rowcolor[rgb]{0.93,1.0,0.87} \textbf{TransPoser} & \checkmark & \checkmark & \checkmark & \checkmark & \checkmark & \footnotesize{DeepDDF} & \footnotesize{Cross-Attn.} \\
        \bottomrule[1.5pt]
    \end{tabular}
    \endgroup
    \caption{Category-level object shape and pose estimation methods. TransPoser realizes neural optimization that effectively weighs multiple viewpoints and leverages learned momentum, which enables efficient and accurate joint object shape and pose estimation from sequential observations. We also introduce an efficient neural shape representation, DeepDDF, for this.}
\label{tab:methods_comparison}
\end{table*}

\section{Introduction}

Rigid object shape and pose estimation from a sequence of visual observations is a core task of computer vision. For instance, a mobile robot in a room would need to constantly recognize and localize objects such as chairs and tables in the environment that can be moved from time to time. A natural sensory input to this would be RGB-D images sequentially acquired as the observer moves around in a scene. Recovery of the complete object shape and its pose would be essential for situational awareness which requires correct anticipation of obstructions by and understanding of the utility of objects in the scene.

A wide variety of methods have been proposed to achieve this joint shape and pose estimation. A representative approach is object SLAM~\cite{mccormac2018fusion++,wu2020eao}, which introduces rigid objects into the simultaneous localization and mapping paradigm so that individual objects can be isolated and localized. These methods either assume that the objects found in the scene are known and their shapes can be pre-acquired (\ie, detection and mapping) or only reconstruct the visible parts of them. Recent works have tackled joint estimation of the complete object shape and its pose from the observed RGB-D image sequence~\cite{wang2021dsp,shan2021ellipsdf}. These methods, however, rely on conventional non-linear optimization, which fundamentally makes sequential online estimation hard.

As shown in \cref{fig:opening}, in this paper, we propose to learn to jointly estimate the shape and pose of objects.
We formulate this joint estimation with neural networks so that the estimation can be achieved with forward inference. This lets us avoid complex non-linear optimization and makes the joint estimation amenable to potential on-board implementation. Most important, it lets us fully leverage the structured bias underlying category-level object pose estimation as a learned estimator.

Our method is based on two key ideas. The first is to represent the shape in the 2D view-space. For this, we introduce Deep Directional Distance Function (DeepDDF), a neural network that outputs the depth image of an object given the camera viewpoint and viewing direction. This enables fast error computation with the observed depth images, which is essential for efficient estimation of the shape and pose. DeepDDF is a category-level model, trained on 3D shapes of the same category and lets us reconstruct the complete shape of a category instance (\eg, a particular chair) from the observed RGB-D images.

The second is to formulate the joint estimation itself as a Transformer~\cite{vaswani2017attention}, which we refer to as TransPoser. We fully leverage the two characteristics of Transformer: tokenization and multi-head attention. Tokenization lets us encode sequentially acquired observations as discrete tokens of variable cardinality. This is essential for realizing an optimization that can handle a variable number of observations as they come in. The attention scheme lets us learn to transform the observations and predicted shapes into an embedding space most suitable for minimizing their discrepancy. Most important, they let us weigh the contribution of each observation and the previous estimates, \ie, they realize a learned weighted optimization with learned momentum.

Our method seamlessly integrates DeepDDF and TransPoser to estimate the shape and pose of an object captured in a sequentially obtained set of RGB-D images. We conduct extensive experiments to evaluate the effectiveness of DeepDDF and TransPoser on both synthetic and real RGB-D data. We compare with the few competing methods that achieve the same task albeit with conventional non-linear optimization. The experimental results show that DeepDDF achieves high accuracy as a category-level object shape representation and TransPoser efficiently and accurately estimates both the shape and pose of those objects. To our knowledge, TransPoser achieves state-of-the-art accuracy on this challenging task. We believe DeepDDF and TransPoser will serve as a sound foundation for effective joint object shape and pose estimation, which has implications in a broad range of areas including robotics, VR/AR, and autonomous driving.

\section{Related Work}

\paragraph{Joint Object Pose and Shape Estimation}
Estimation of object shape and its pose from a set of visual observations, most often captured sequentially by a moving observer (\eg, mobile robot), is a fundamental task for visual scene understanding. Most approaches to this task assume RGB or RGB-D video as input. Online methods, such as object-aware SLAM~\cite{mccormac2018fusion++,wu2020eao}, have also been introduced but none of them recover the complete object shape and instead result in crude approximations or partial reconstructions.

These past works naturally rely on iterative non-linear optimization, which necessitates repeated evaluation of the prediction error between the hypothesized object shape with predicted pose and the observations~\cite{Sucaretal3DV2020,wang2021dsp,bruns2022sdfest,deng2022icaps,runz2020frodo,maninis2022vid2cad}. Vid2CAD~\cite{maninis2022vid2cad}, for instance, projects a CAD model to the multi-view observations to compute the loss for pose update. Other methods~\cite{wang2021dsp,shan2021ellipsdf} employ neural 3D shape representations such as DeepSDF~\cite{park2019deepsdf} to represent unknown shapes. ELLIPSDF~\cite{shan2021ellipsdf}, for instance, adopts a bi-level object shape model consisting of an ellipse and an SDF for estimation from point clouds observed as RGB-D videos.

These methods, however, fundamentally suffer from high computational cost mainly due to their reliance on conventional non-linear optimization. Conventional numerical non-linear optimization requires numerous steps of updates that also limit their ability to adapt to sequentially incoming observations. The shape representations adopted by these studies also demand high computational cost for prediction error evaluation as 2D views must be explicitly rendered after the 3D representation is instantiated. Numerous iterations of this causes prohibitive cost for any possibility of online computation. We tackle joint object shape and pose reconstruction by performing optimization sequentially under a practical setting where new observations are added online.  As shown in~\cref{tab:methods_comparison}, we leverage a data-driven, efficient, and accurate neural optimization and neural shape representation which makes the estimation just a handful of forward inference passes.

%%%

\vspace{-12pt}
\paragraph{Neural Shape Representation}
An efficient and accurate object shape representation is essential for iterative optimization for shape estimation as it dictates the computational cost of a single iteration. Commonly used representations such as point clouds~\cite{tian2020shape,wang2021category,Chen_2021_CVPR,chen2021sgpa,lin2022category} and  voxels~\cite{Sucaretal3DV2020,bruns2022sdfest} are inefficient as they first need to be instantiated in 3D before their 2D depth images can be rendered. Recently, implicit neural representations (INRs) have been introduced as a continuous but compact shape representation~\cite{bruns2022sdfest,runz2020frodo,shan2021ellipsdf}. Typical examples of INRs, such as DeepSDF~\cite{park2019deepsdf} and Occupancy Networks~\cite{mescheder2019occupancy}, have been used in conjunction with optimization pipelines~\cite{runz2020frodo,deng2022icaps,shan2021ellipsdf}, but they still require substantial computational and memory costs for rendering. The proposed DeepDDF directly outputs depth images from its latent representation. In contrast to similar approaches realizing direct shape evaluation~\cite{zobeidi2021deep,aumentado-armstrong2022representing,yenamandra2022hdsdf}, the key novelty of our neural shape representation lies in the enforcement of multi-view consistency with a low-cost voxel representation placed before the depth image decoder. We believe this idea of combining low-res 3D and high-res 2D in a single neural representation will have implications in applications beyond joint shape and pose estimation.

\vspace{-12pt}
\paragraph{Learning-based Optimization}
Learning-based optimization can efficiently optimize intractable cost functions in a data-driven fashion~\cite{denil2016learning,adler2017solving,clark2017meta,li2016learning}. Numerous methods have been introduced for learning to optimize 2D alignment problems~\cite{von2020gn,clark2018ls,lin2017inverse} or simple optimization benchmarks~\cite{denil2016learning,gartner2022transformer}. Its application to object pose estimation is, however, relatively underexplored. Notably, Ma \etal~\cite{ma2020deep} tackled the ill-posed problem of estimating object pose from monocular 2D observations with learnable optimization networks. Their work is, however, limited to single-image inputs and uses simple LeNet layers that directly regress $\Delta \bm x$ based on the current estimates. In contrast, our method extends a Transformer architecture to learn a challenging, multi-view optimization problem using learnable momentum, which allows us to robustly and efficiently optimize the complex objective by tapping into the structured bias of category-level shape and pose estimation.

TransPoser generalizes these neural optimization approaches in two fundamental ways.  It infers the update based not only on the current estimate but also on those of past iterations. It also takes in multiple observations from different views as a discrete input sequence. This enables iterative optimization with learned weights on the multi-view images and achieves this with a learned momentum that fully leverages the structured bias in optimization of joint shape and pose estimation at the category level.

\section{Joint Shape and Pose Estimation}

Our goal is to jointly estimate the complete shape and pose of a rigid object observed in a sequentially growing set of RGB-D images captured as the viewpoint and viewing direction changes. The key challenges underlying this task is the reconstruction of the complete shape of an object from its partial depth observations each captured from a different viewing direction and the estimation of object pose in a canonical coordinate frame. We define the canonical coordinate frame for each object category (object coordinate frame) and estimate the object pose as the 3D translation, 3D rotation, and 3D scale from it in the first observed depth image (world coordinate frame). The viewpoint and viewing directions for each RGB-D observation with respect to those of the first depth image can be computed from background features (\eg, regular SLAM). As such, the relative observer motion can be safely assumed to be known.

Jointly solving for the shape reconstruction and pose estimation naturally necessitates an optimization that evaluates the discrepancy between the hypothesized object shape in a predicted pose. This requires repeated rendering of a depth image of the predicted object which can easily become the bottleneck of the joint estimation. Conventional 3D object representations including mesh models, signed distance functions, and occupancy volumes, regardless of whether they can be evaluated with forward inference on a network, add significant burden to this process as the depth image needs to be rendered from the instantiated 3D object.

We instead introduce a novel shape representation that directly outputs a depth image of the object for a given viewpoint and viewing direction. The proposed DeepDDF is inherently a 2D view-based shape representation that bypasses the need for 3D shape rendering. The output depth image can be directly compared with the observed depth image, which significantly cuts down the computational burden of the joint shape and pose estimation.

The joint estimation is a highly complex non-linear optimization whose parameters consist of the 3D shape encoded with the DeepDDF as a low-dimensional latent code and also the 9 degrees-of-freedom (DoF) pose parameters. Notably, we estimate the 3D scale together with the usual 6-DoF pose encoding the 3D translation and 3D rotation of the object. The challenge also lies in the fact that we would like to fully leverage the observations which are obtained sequentially by gradually updating the parameters as the new observations come in.

This necessitates an efficient optimizer that can take in a variable sequence of input RGB-D observations and that can dynamically weigh the contributions of the sequentially added observations. We achieve this by deriving a novel Transformer, which we refer to as TransPoser, that directly outputs the updates to the shape and pose parameters for a given partial sequence of observed RGB-D images. Once learned, one step of the optimization is a single forward inference, which dramatically cuts down the computational cost compared to conventional non-linear optimization based joint estimation that needs to restart the iteration every time a new observation is captured.

\begin{figure}[t]
\centering
  \includegraphics[width=0.85\linewidth]{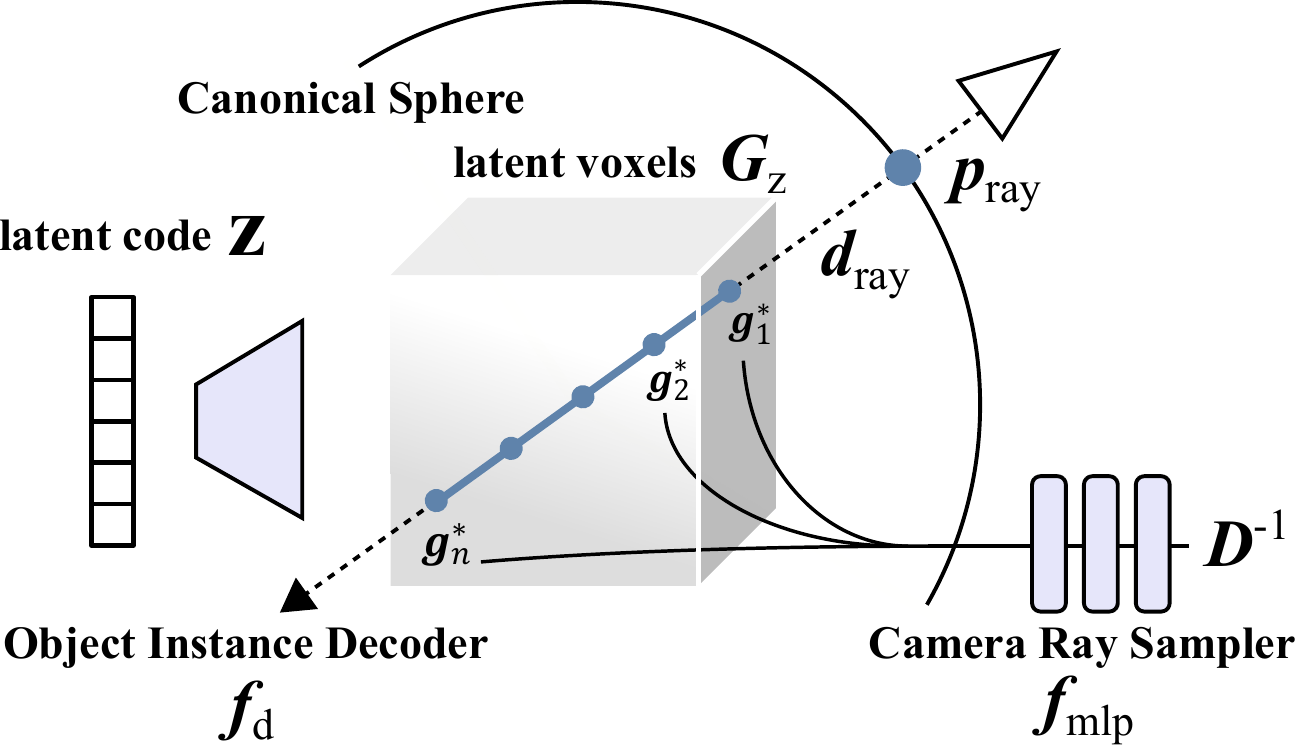}
\caption{DeepDDF returns the inverse distance to a object surface it represents for a given latent code, viewpoint, and viewing direction. It leverages a low-resolution 3D latent volume, which is decoded from the latent code with the Object Instance Decoder, and samples rays in this volume for a give viewpoint and viewing direction using the Camera Ray Sampler. These decoder and sampler collectively enable efficient and multi-view consistent 3D representation with direct 2D image-space output.}
\label{fig:deepddf}
\end{figure}

\section{Deep Directional Distance Function}
We introduce Deep Directional Distance Function (DeepDDF) as a novel 3D representation of rigid objects. Similar to DeepSDF \cite{park2019deepsdf}, DeepDDF is a decoder-based model trained on category-level object instances. Each object category is represented with one DeepDDF. In sharp contrast to  other 3D object representations including DeepSDF, DeepDDF directly outputs a depth image from a latent code encoding the object instance by conditioning on a given viewpoint and viewing direction. For each camera ray $\vct{d}_\text{ray}$ from a viewpoint $\vct{p}_\text{ray}$, DeepDDF returns the distance $D$ from the camera
\begin{equation}
D = f_{\text{DDF}}(\vct{p}_\text{ray}, \vct{d}_\text{ray}, \vct{z})\,,
\end{equation}
where $\vct{z}$ is the latent code of the object instance.
\begin{figure*}[t]
  \centering
  \includegraphics[width=\linewidth]{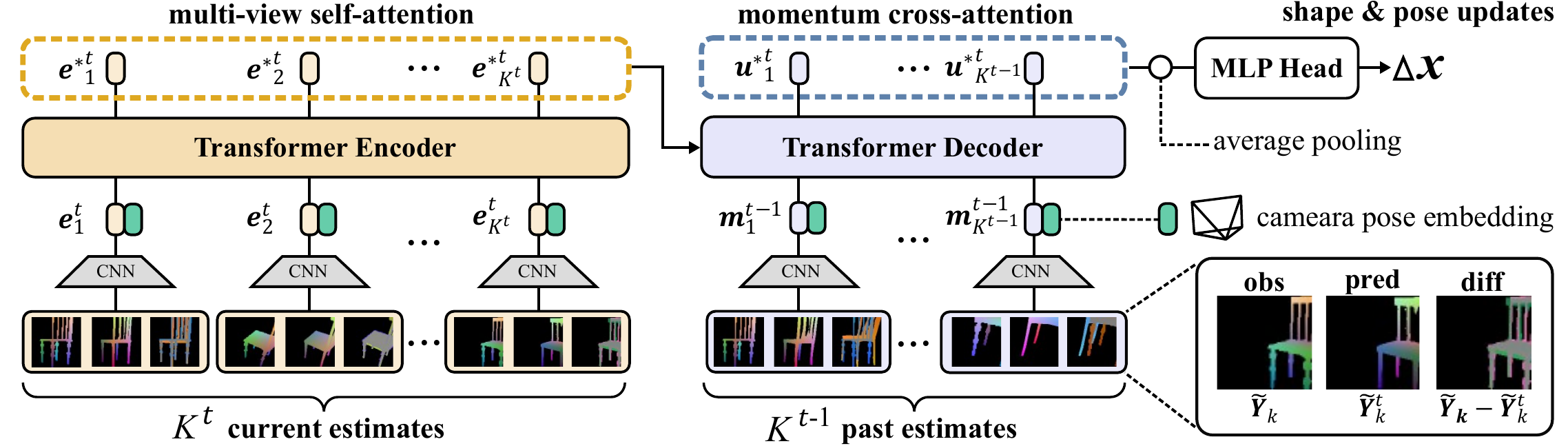}
  \vspace{-21pt}\\
  \caption{TransPoser realizes neural optimization with a Transfomer encoder-decoder architecture. The multi-view sequential observations are tokenized and cross-attended with queries representing past estimates and current view, which enables a learned weighted optimization with learned momentum in iterated forward inference passes.}
  \label{fig:optnet}
\end{figure*}

DeepDDF enables direct evaluation of the object shape and pose against observations in the image-space as it bypasses rendering (projection) of 3D geometry in the 2D view. Two fundamental challenges, however, need to be resolved to achieve this direct 2D view-dependent representation. The first is that the depth values for each camera ray across different viewpoints must be consistent, \ie, must represent the same rigid object shape. The second is the large number of samples required to span the entire space of the inputs to the decoder $\vct{p}_\text{ray}$, $\vct{d}_\text{ray}$, and $\vct{z}$.

We overcome these challenges by introducing a sparse 3D voxel grid aligned with the target object space.
Inspired by DeepVoxels~\cite{sitzmann2019deepvoxels}, which improves the accuracy of RGB novel view synthesis with voxel features, we introduce a 3D voxel representation of the rough object shape before the decoder which ensures multi-view consistency of output depth images. This 3D voxel can be of very low-resolution as it does not encode the 3D shape itself.

As shown in \cref{fig:deepddf}, the latent code $\vct{z}$ is decoded into latent voxels by an object instance decoder $\fd$, and the voxels on the camera rays, defined by its 3D viewpoint $\vct{p}_\text{ray}$ and 3D direction $\vct{d}_\text{ray}$, are sampled by a camera ray sampler $\fmlp$. This low-resolution but three-dimensional voxel corresponding to the object instance encoded by the latent code $\vct{z}$ enforces consistency across the camera rays and makes the output depth values $D$ multi-view consistent.  It also enables reduction in the number of samples for $\vct{p}_\text{ray}$, $\vct{d}_\text{ray}$, and $\vct{z}$ by ensuring this multi-view consistency.

We implement $\fd$ with a 3D deconvolution network and $\fmlp$ with an MLP. Please see the supplemental material for the full network architecture. The object instance decoder $\fd$ decodes the latent code $\vct{z}$ into a 3D voxel array $G_z$ encaging the target shapes. Each voxel $\vct{g}$ of $G_z$ stores a feature vector encoding the 3D shape corresponding to $\vct{z}$. The camera ray sampler $\fmlp$ regresses the inverse distance from a viewpoint at a unit distance to the object surface
\begin{equation}
D^{-1}  = f_{\text{mlp}}\left(\bigoplus [\vct{g}^*_1, \vct{g}^*_2, ..., \vct{g}^*_{N}]\right) \,,
\end{equation}
where $\bigoplus [\vct{g}^*_1, \vct{g}^*_2, ..., \vct{g}^*_{N}]$ denotes the concatenated feature vectors of $N$ voxels $\{\vct{g}^*_i\}^{N}_{i=1}$ sampled along the camera ray represented by $\vct{p}_\text{ray}$ and $\vct{d}_\text{ray}$.  Regressing the inverse distance prevents $\fmlp$ from returning infinity for rays outside the object region.

We train DeepDDF by optimizing the latent code $\vct{z}$ and the network parameters~\cite{park2019deepsdf} by minimizing the loss between the predicted depth and surface normal from the ground truth
\begin{align}
\mathcal L_{\text{DDF}} &= \mathcal L_\text{dst}+\lambda_{\text{nrm}} \mathcal L_{\text{nrm}} + \lambda_{\text{ltn}} ||\vct{z}|| \,,\\
\label{eq:ddf_loss}
\mathcal L_\text{dst} &= \left| D^{-1}_{\text{est}} -  D^{-1}_{\text{gt}}\right| \,, \\
\mathcal L_\text{nrm} &= \text{MAE}(\vct{n}_{\text{est}}, \vct{n}_{\text{gt}}) \,,
\end{align}
where $\mathcal L_\text{dst}$ is the discrepancy between the current depth $D_\text{est}$ and ground truth $D_\text{gt}$, and $\mathcal L_\text{nrm}$ is the mean absolute error of the estimated and ground truth normal directions, $\vct{n}_{\text{est}}$ and $\vct{n}_{\text{gt}}$, respectively. We compute the normal by sampling the depth of the neighboring rays. The latent codes $\vct{z}$ are regularized with the term $||\vct{z}||$ and $\lambda_{\text{nrm}}$ and $\lambda_{\text{ltn}}$ control the weighting between the three terms which are determined empirically. We used $1.0\times10^{-2}$ and $1.0\times10^{-4}$, respectively, in all experiments.

As shown in \cref{fig:deepddf}, we avoid redundant sampling by training the network with viewpoints on a fixed-radius sphere which we refer to as the canonical sphere. In our implementation, we set the radius to the diagonal distance of the bounding box of a given object category, which we normalize to 1, and its center at the origin of the object coordinate frame. When generating the depth for viewpoints inside or outside this sphere, we use the intersection of the camera ray with this sphere and add the distance to this sphere to the depth for the given point and direction on the canonical sphere.

%------------------------------------------------------------------------

\section{TransPoser}

Let $\vct{x}$ denote a vector of parameters of joint shape and pose estimation: the object shape as the latent code of DeepDDF $\vct{z}$ and its 9D pose of 3D translation, 3D rotation, and 3D scaling. Note that the 3D shape represented by DeepDDF is internally normalized to have a unit diagonal length for its 3D bounding box. The 3D scaling of DeepDDF is thus relative to this diagonal length, which increases the representation power of DeepDDF. The 3D scaling estimated for pose then further aligns the predicted 3D shape to the observations. This cascaded scaling estimation helps express a variety of 3D shapes and their accurate poses.

Given the current estimate $\vct{x}^t$ at step $t$, the goal of TransPoser is to return the update $\Delta \vct{x}^t$ that refines the shape and pose $\vct{x}^{t+1} = \vct{x}^t + \Delta \vct{x}^t$ computed from the multi-view observed RGB-D images. We denote the number of images at step $t$ with $K^t$. As depicted in \cref{fig:optnet}, TransPoser is an encoder-decoder Transformer~\cite{vaswani2017attention} and uses DeepDDF to synthesize the depth image for each $\vct{x}$. Each of the tokens to the TransPoser encoder is generated from the observed depth image for the $k-$th view $\vct{Y}_k$ and the corresponding predicted depth image for the $t-$th step $\vct{Y}_k^t$.

The depth images $\vct{Y} \in \mathbb{R}^{H \times W \times 1}$ are first converted to images that represent object surface coordinates, which we refer to as surface maps $\widetilde{\vct{Y}} \in \mathbb{R}^{H \times W \times 3}$. Instead of the depth to the surface in $\vct{Y}$, each pixel of $\widetilde{\vct{Y}}$ stores the 3D location in the normalized object coordinate space (NOCS)~\cite{wang2019normalized} representing the same position on the surface. We also assume we have a mask for the object region, which usually can be supplied through semantic segmentation. The pixels outside this mask are all zeroed out in the observation. The predicted images, and thus their surface maps, also have zero values rendered for the background.

This use of a surface map representation is one of the key novel ideas underlying TransPoser. By consolidating the observed and predicted depth values for their individual camera coordinates into the same object coordinate frame, but in a spatial image, TransPoser is able to compute self- and cross-attention without having to overcome the differences in coordinate frames. It also avoids ``projection breakdown'' caused by 2D convolutions on depth images~\cite{jiang2022uni6d}.

We compute 3 surface maps from the 2 depth images, namely the observed surface map, predicted surface map, and their difference. The last encodes the discrepancy in the surface point coordinates for each pixel by simply taking the difference between the first two. By explicitly preparing this difference surface map, we lessen the burden on the network to learn to evaluate the prediction error. These 3 surface maps, each a 3 channel image of surface coordinates, are input to a backbone feature extraction network independently as a 4 channel image with the last filled with the object mask. The feature maps are concatenated and passed through a fully-connected layer to produce the corresponding token denoted as $\vct{e}^t_k$ for the encoder and $\vct{m}^{t-1/t}_k$ for the decoder. We use ResNet-18~\cite{he2016deep} as the backbone feature extractor for both the encoder and decoder. In addition, we also encode the viewing condition of each token by adding a 7D vector consisting of the 3D camera viewpoint in the object coordinate frame, the 3D camera ray direction towards the center of the object bounding box, and the ratio between the diagonal of the entire image to the diagonal of the bounding box. Note that we do not use regular positional encodings~\cite{vaswani2017attention}. The observations are image sets, even though they are sequentially acquired, and we explicitly encode their viewing conditions but not their order.

Self-attention in the encoder transforms each view and associated prediction and error while consolidating information across the views. This effectively learns the weighting between the views. Consider, for example, estimating the pose of a chair. The optimization would ideally put more weight on the prediction errors on views that capture both the backrest and seat, compared to views only capturing the handrail or the legs. The encoder computes self-attention between tokens $\{\vct{e}_k^t\}_{k=1}^{K^t}$ of multiple $K^t$ views for the current estimate $\vct{x}^t$. The encoder has three layers of self-attention and the outputs are denoted with $\{{\vct{e}^{\ast}}_k^t\}_{k=1}^{K^t}$.

\begin{figure}[t]
    \centering
    \includegraphics[width=\linewidth,trim=108 298 500 30,clip,page=1]{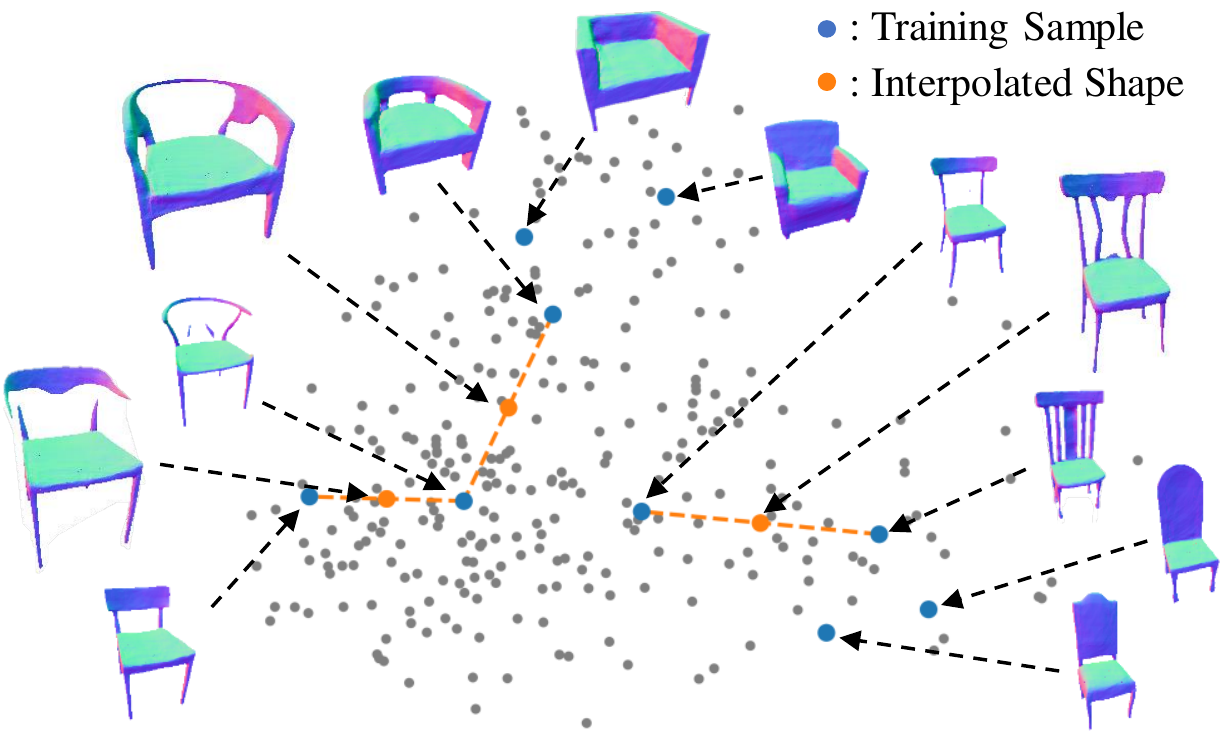}
    \caption{The 2D PCA subspace spanned by the learned latent codes of DeepDDF. It consists of clusters of similar objects and linear interpolation in the space creates objects with shape features inherited from end points. These results show that DeepDDF successfully extracts the structured coherence of the shape space of the object category.}
    \label{fig:ddf_pca}
\end{figure}

The decoder query tokens are each computed from the same combination of observed, predicted, and discrepancy depth images. Unlike the encoder tokens, they represent those of different time steps. As depicted in \cref{fig:optnet}, our key idea is to formulate the decoder to learn a momentum-guided update from the self-attended current estimates $\{\vct{e}^{\ast t}_k\}_{k=1}^{K^t}$ (\ie, encoder outputs) and past estimates $\{\vct{m}^{t-1}_k\}_{k=1}^{K^{t-1}}$ and also the current view $\vct{m}^t_{K^t}$ (\ie, decoder queries), if the camera moved. When the camera moves between the two optimization steps $t-1$ and $t$, $K^t$ and $K^{t-1}$ are not the same and we simply use the current estimates $\vct{x}^t$ as viewing condition encoding of the latest view $\vct{m}^t_{K^t}$. The decoder has three layers of self-attention and three layers of cross-attention. The self-attention layers effectively weigh multi-view contributions of the past estimates similar to the self-attention in the encoder. Most important, the cross-attention layers weight contributions across the current and past estimates which effectively amounts to a learned momentum term to output the decoder tokens $\{\vct{u}^{\ast t}_k\}$. These output tokens are average-pooled, passed through an MLP and converted back into the world coordinate frame (the first view) to finalize the additive update $\Delta\vct{x}^t$ to the current estimate $\vct{x}^t$.

TransPoser does not have layer normalizations which are common in regular Transformers~\cite{vaswani2017attention}. This is because layer normalization corresponds to normalizing the gradient vector in regular SGD and would cause the network to ignore error minimization. Omitting layer normalization, however, has as a side-effect of making the training unstable. We use T-Fixup~\cite{huang2020improving} initialization to resolve this.

The loss of TransPoser is the component-wise error between the updated estimate $\vct{x}^{t}+\Delta \vct{x}^t$ and the ground truth values $\vct{x}_\text{gt}$, similar to that of Ma \etal. The details of update formulae and the loss function for each parameter value are described in the supp\onedot material. Unlike Ma \etal, however, we use one common network among all iterations. We update the network parameters after each step $t$.

\begin{table}[t]
    \centering
    \begingroup
    \setlength{\tabcolsep}{1.5mm}
    \begin{tabular}{l||cc|cc}
        \toprule[1pt]
        \multirow{2}{*}{Model}
        & \multicolumn{2}{c|}{Comp. Cost} & \multicolumn{2}{c}{Shape Accuracy} \\
         & Time & Memory & Chair & Table \\
        \hline
        DIST \cite{liu2020dist} & 0.27 sec & 5.4 GB & 1.28 & 1.43 \\
        \rowcolor[rgb]{0.93,1.0,0.87}  \textbf{DeepDDF} & \textbf{0.04} sec & \textbf{5.0} GB & \textbf{0.83} & \textbf{0.91} \\
        \bottomrule[1pt]
    \end{tabular}
    \vspace{-0.5\baselineskip}
    \endgroup
    \caption{Experimental comparison of shape reconstruction speed and accuracy of DeepDDF with DIST. Accuracy is shown in Chamfer distances $(10^{-3})$ from ground truth. Our DeepDDF realizes more accurate reconstruction in an order of magnitude less time.}
    \label{tab:ddf_quant}
\end{table}
\begin{figure}
    \centering
        \setlength{\tabcolsep}{0mm}
        \begin{tabular}{cccccccc}
            \; &
            \footnotesize{Obs.} &
            \footnotesize{DIST \cite{liu2020dist}} &
            \multicolumn{1}{c}{\footnotesize{DeepDDF}} &
            \; &
            \footnotesize{Obs.} &
            \footnotesize{DIST \cite{liu2020dist}} &
            \footnotesize{DeepDDF} \\
             &  &  &  &  &  &  & \vspace{-15pt}\\
             &  &  & \multicolumn{1}{c}{} &  &  &  & \vspace{-10pt}\\
            \; &
            \includegraphics[height=1.37cm, trim = 0 -20 0 -20, clip]{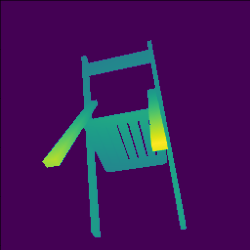} &
            \includegraphics[height=1.37cm, trim = 5 0 0 0, clip]{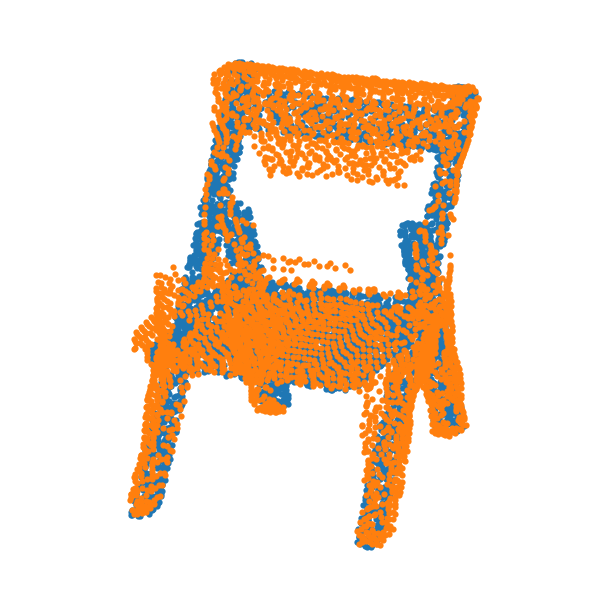} &
            \multicolumn{1}{c}{\includegraphics[height=1.37cm, trim = 5 0 0 0, clip]{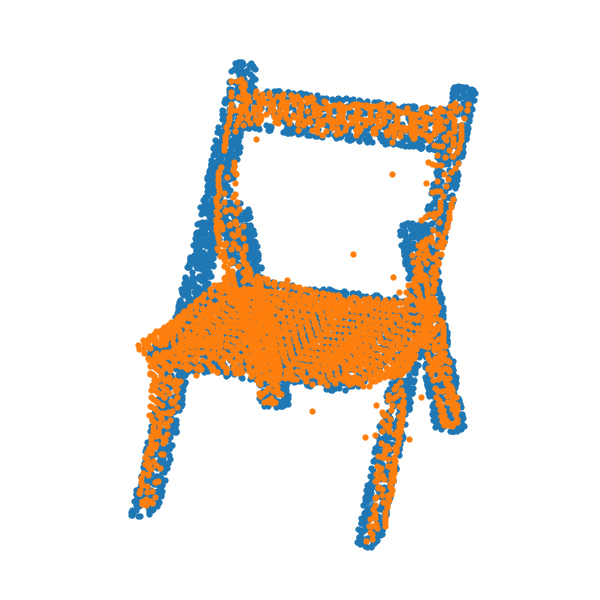}} &
            \; &
            \includegraphics[height=1.37cm, trim = 0 -20 0 -20, clip]{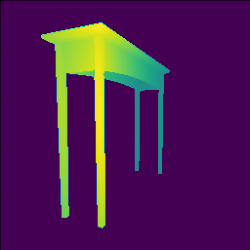} &
            \includegraphics[height=1.37cm, trim = -12 10 0 20, clip]{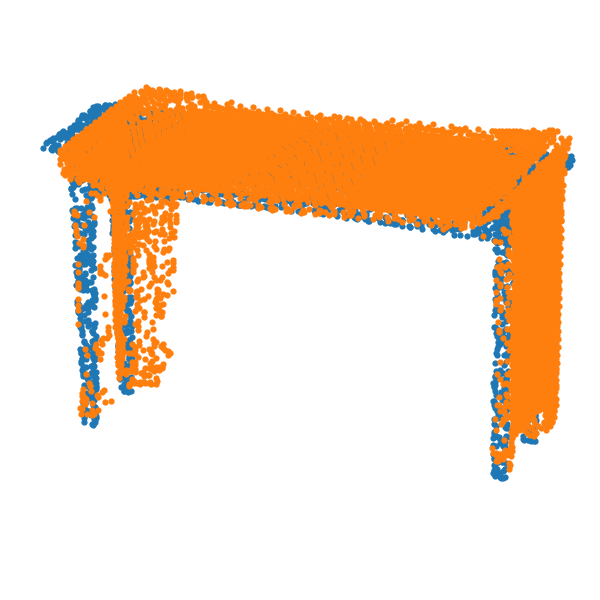} &
            \includegraphics[height=1.37cm, trim = -12 10 0 20, clip]{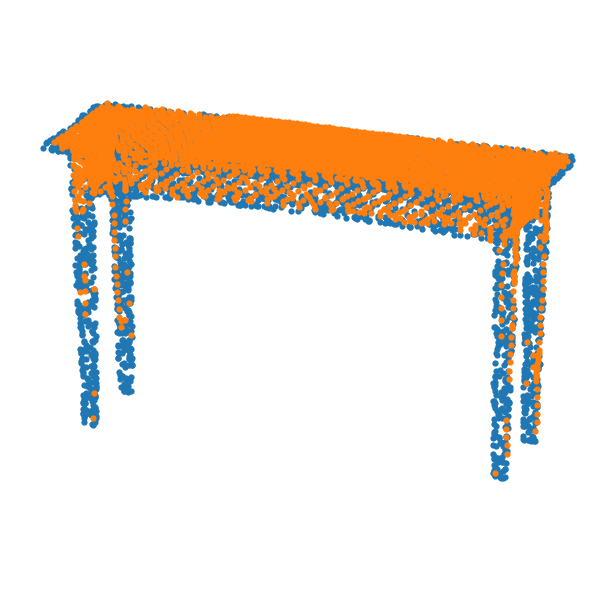}\\
             &  &  &  &  &  &  & \vspace{-17pt}\\
             &  &  & \multicolumn{5}{r}{\includegraphics[width=3.6cm, trim = 0 0 0 0, clip]{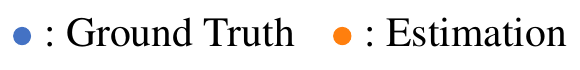}}\\
             &  &  &  &  &  &  & \vspace{-23pt}\\
        \end{tabular}
\caption{Example shape reconstruction with conventional optimization using DIST and DeepDDF. The results of DeepDDF are qualitatively more accurate.}
\label{tab:ddf_quali}
\end{figure}

\begin{table*}[t]
    \centering
    \begingroup
    \setlength{\tabcolsep}{1.5mm}
    \begin{tabular}{l||ccc|cccccccccccc}
        \toprule[1.5pt]
        \multirow{2}{*}{Model} &
        \multicolumn{3}{c|}{Cls avg.} & \multicolumn{3}{c}{Cabinet} & \multicolumn{3}{c}{Chair} & \multicolumn{3}{c}{Display} & \multicolumn{3}{c}{Table} \\
         & T & R & S & T & R & S & T & R & S & T & R & S & T & R & S \\
        \hline
        Baseline based on~\cite{ma2020deep} & 1.58 & 1.70 & 4.24 & 1.98 & 1.56 & 4.97 & 1.17 & 0.60 & 3.17 & 1.72 & 1.05 & 3.03 & \textbf{1.46} & 3.59 & 5.79 \\
        w/o Dec.                   & 1.45 & 1.95 & 3.55 & 1.45 & 1.54 & \textbf{3.56} & 1.17 & 0.55 & \textbf{2.59} & 1.49 & 1.17 & 2.44 & 1.70 & 4.55 & 5.60 \\
        \rowcolor[rgb]{0.93,1.0,0.87}  TransPoser                 & \textbf{1.38} & \textbf{1.44} & \textbf{3.43} & \textbf{1.15} & \textbf{1.40} & 3.59 & \textbf{0.96} & \textbf{0.54} & 2.68 & \textbf{1.39} & \textbf{1.02} & \textbf{2.34} & 2.02 & \textbf{2.82} & \textbf{5.09} \\
        \bottomrule[1.5pt]
    \end{tabular}
    \vspace{-0.5\baselineskip}
    \endgroup
    \caption{
    Ablation study of TransPoser architecture using synthetic data. The baseline \cite{ma2020deep} corresponds to an architecture without multi-view and momentum. ``w/o Dec.'' stands for TransPoser without the decoder, \ie, without learned momentum.  Metrics for translations (T), rotations (R), and shapes (S) are mean squared error $(10^{-2})$, Riemannian distance $(10^{-1})$, and Chamfer distance $(10^{-3})$, respectively.
    For almost all categories, the proposed TransPoser, which learns to weigh multi-view observations and to leverage learnable momentum, achieves highest accuracy.
    }
    \label{tab:dfnet_ablation}
\end{table*}
%------------------------------------------------------------------------
\section{Experimental Results}
We thoroughly evaluate the effectiveness of DeepDDF and TransPoser with large-scale synthetic and real datasets.  Please refer to the supp\onedot material for details on the evaluation setup and additional results.

\subsection{DeepDDF}
We train one DeepDDF network for each category using CAD models from ShapeNet~\cite{chang2015shapenet}. As training data, we rendered $512\times 512\text{px}$ depth images from virtual cameras of $60^\circ$ field-of-view, located randomly on the unit sphere centered at the target model. We synthesized 200 pairs of depth and normal maps for each of 1,465, 3,196, 986, and 4,817  CAD models in the \textit{Cabinet}, \textit{Chair}, \textit{Display}, and \textit{Table} categories, respectively. We used the same split of the \textit{Chair} and \textit{Table} categories as DeepSDF~\cite{park2019deepsdf} except for several CAD models with transparent parts or inverted normals.

\Cref{fig:ddf_pca} visualizes the 2D PCA subspace spanned by the latent codes of DeepDDF for \textit{Chair}. The latent space contains clusters of similar shapes in the same object subcategory, such as single sofas, and interpolation synthesizes novel shapes inheriting the features of the end points. This demonstrates that DeepDDF is able to extract a structured latent space of object instances in the same category.

We compare the accuracy and efficiency of DeepDDF with DIST~\cite{liu2020dist} on shape reconstruction. DIST is also a single image shape estimator, but uses DeepSDF with iterative sphere tracing to render the depth image. Unfortunately, we cannot compare with past methods that directly output distance~\cite{zobeidi2021deep,aumentado-armstrong2022representing,yenamandra2022hdsdf} as their original implementations were not available. We use the \textit{Chair} and \textit{Table} categories which contain challenging geometric structures like thin legs and has a wide variety in part features. We assume that both the camera and object poses are known, and optimize the latent code to minimize the error between an observed and a synthesized depth image with regular non-linear optimization with Adam optimizer to directly compare the two in their effectiveness purely in shape representation.

\Cref{tab:ddf_quant} shows quantitative results of this experimental comparison, in computational cost and reconstruction accuracy. We evaluate memory usage and inference times to generate an $256\times 256\text{px}$ image for computational costs, and the Chamfer distance between the ground truth and the estimated point clouds for reconstruction accuracy. The results show that our DeepDDF is superior to DIST in both aspects. DeepDDF achieves a magnitude of faster estimation thanks to its direct depth image output. \Cref{tab:ddf_quali} shows reconstructed shapes. These results show that DeepDDF also reconstructs qualitatively more accurate shapes.

\subsection{TransPoser}
We evaluate the effectiveness of TransPoser for joint shape and pose estimation on synthetic data, and also compare it with past methods on ScanNet~\cite{dai2017scannet} which is a dataset of scanned real scenes and objects.

\vspace{-12pt}
\paragraph{Ablation Studies on Synthetic Data}
To train TransPoser, we randomly generate camera trajectories moving around the object, and sample 5 viewpoints along each of them to render $128\times 128\text{px}$ depth images of $50^\circ$ field-of-view. TransPoser is trained to optimize the shape and pose by 10 iterations during which the number of observations increases at every 2 iterations.  The pose is initialized randomly, and the shape is initialized by the mean latent code of the shape category.

We train one TransPoser network for each category using CAD models from ShapeNet. DeepDDF is pre-trained, and frozen during the training of TransPoser.
We compare TransPoser with the following models that each correspond to ablation of key architectural components. The ablated models are a baseline model composed of a backbone network and an MLP head, and TransPoser without the decoder. The baseline model is based on Ma~\etal~\cite{ma2020deep}, which does not exploit the relative postures of the multiple viewpoints nor consider the momentum during the optimization. This model inputs each feature $\vct{e}_k^t$ from the backbone network to the MLP head, calculates the updated value $\Delta \vct{x}_k^t$ for each viewpoint $k$ independently, and then takes their average as $\Delta \vct{x}^t$. TransPoser without the decoder only weighs the multiple viewpoints. This model averages the encoder outputs $\{{\vct{e}^{\ast}}_k^t\}_{k=1}^{K^t}$, and inputs it to the MLP head.

We evaluate the errors in translation, rotation, and shape estimation by the mean squared error, the Riemannian distance~\cite{moakher2002means}, and the bi-directional Chamfer distance normalized by the object scale, respectively.  On computing the shape error by Chamfer distance, we generated a point cloud corresponding to the estimated latent code by integrating DeepDDF depth-maps rendered from 5 viewpoints distributed uniformly around the object.

\Cref{tab:dfnet_ablation} shows the estimation errors for each category.
These results demonstrate quantitatively that the proposed TransPoser achieves the highest accuracy in most of the metrics, while achieving a faster convergence. TransPoser without the decoder also converges as fast as TransPoser, but gets trapped in local minima as shown in the rotation error. This implies that referencing the past optimization steps through the learned cross-attention by the decoder (\ie, the learned momentum) helps converge to a better minima.

\vspace{-12pt}
\paragraph{Real-world Data}

\begin{table}[t]
    \centering
    \begingroup
    %\small
    \setlength{\tabcolsep}{1.5mm}
    \begin{tabular}{l||c|cccc}
        \toprule[1.5pt]
        Method & \footnotesize{\!Cls Avg.\!} & \footnotesize{Cabinet} & \footnotesize{Chair} & \footnotesize{Display} & \footnotesize{Table} \\
        \hline
        ELLIPSDF \cite{shan2021ellipsdf} & 89.0 & 91.0 & 90.6 & \textbf{96.9} & 77.3 \\
        \rowcolor[rgb]{0.93,1.0,0.87}  TransPoser & \textbf{94.3} & \textbf{92.4} & \textbf{98.5} & 93.8 & \textbf{92.4} \\
        \bottomrule[1.5pt]
    \end{tabular}
    \vspace{-0.5\baselineskip}
    \endgroup
    \caption{Accuracy of shape estimation for ScanNet~\cite{dai2017scannet}. TransPoser achieves highest accuracy for all categories.}
    \label{tab:scan2cad_shape}
\end{table}
We evaluate the accuracy and efficiency of TransPoser on ScanNet dataset~\cite{dai2017scannet}, which consists of RGB-D videos of real-world indoor scenes. Following ELLIPSDF~\cite{shan2021ellipsdf}, we use the ground-truth object pose, shape, and mask as well as the camera poses. The ground-truth object poses and shapes are provided by Scan2CAD~\cite{Avetisyan_2019_CVPR}.
\begin{table}[t]
    \centering
    \begingroup
    \setlength{\tabcolsep}{1.5mm}
    \begin{tabular}{l|c||c|cccc}
        \toprule[1.5pt]
        Method & \rotatebox{90}{Online} & \rotatebox{90}{Cls avg.\:} & \rotatebox{90}{Cabinet} & \rotatebox{90}{Chair} & \rotatebox{90}{Display} & \rotatebox{90}{Table} \\
        \hline
        Scan2CAD \cite{Avetisyan_2019_CVPR} &  & 31.7 & 34.0 & 44.3 & 17.9 & 30.7 \\
        RayTran \cite{tyszkiewicz2022raytran} &  & 42.1 & 36.2 & 59.3 & 30.4 & 42.5 \\
        \hline
        Vid2CAD \cite{maninis2022vid2cad} & \checkmark & 38.8 & 23.8 & 64.6 & 37.7 & 28.9 \\
        ELLIPSDF \cite{shan2021ellipsdf} & \checkmark & 39.6 & - & - & - & - \\
        \rowcolor[rgb]{0.93,1.0,0.87}  TransPoser & \checkmark & \textbf{59.3} & \textbf{45.4} & \textbf{91.2} & \textbf{45.0} & \textbf{55.7} \\
        \bottomrule[1.5pt]
    \end{tabular}
    \vspace{-0.5\baselineskip}
    \endgroup
    \caption{Accuracy of pose estimation for ScanNet~\cite{dai2017scannet}. TransPoser achieves higher accuracy for all categories.}
    \label{tab:scan2cad_pose}
\end{table}

TransPoser is trained to optimize the shape and pose by 10 iterations, by fine-tuning the model pre-trained with the synthetic dataset.  On fine-tuning, we trained TransPoser using randomly selected 5 frames capturing the object from ScanNet. As the camera moves every \SI{50}{\cm}, we select a frame from the segment such that it captures the target as large as possible, and add it as the new observation.

\begin{figure}
    \centering
    \setlength{\tabcolsep}{0mm}
    \begin{tabular}{ccc}
        \vspace{-13pt} & & \\
        \footnotesize{Observations} & \footnotesize{GT} & \footnotesize{Est.} \\
        \includegraphics[width=5.5cm, trim = 0 0 0 0, clip]{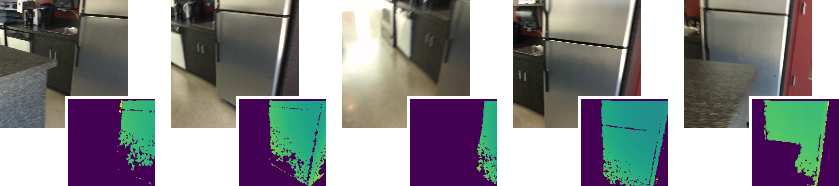} &
        \includegraphics[width=1.3cm, trim = 0 0 0 0, clip]{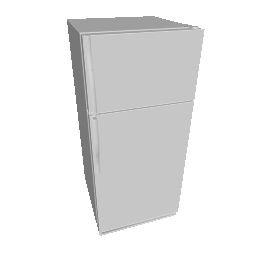} &
        \includegraphics[width=1.3cm, trim = 0 0 0 0, clip]{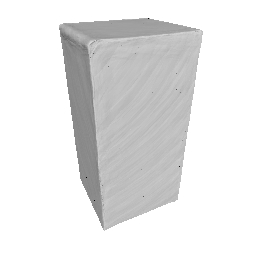} \\
        \includegraphics[width=5.5cm, trim = 0 0 0 0, clip]{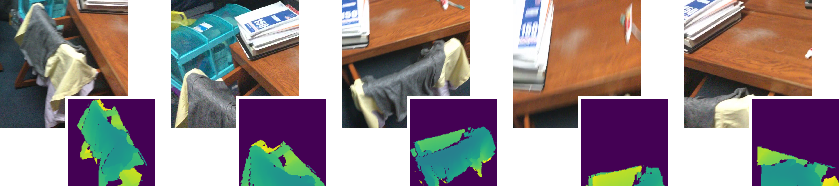} &
        \includegraphics[width=1.3cm, trim = 0 0 0 0, clip]{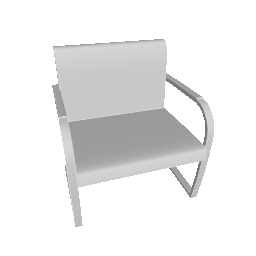} &
        \includegraphics[width=1.3cm, trim = 0 0 0 0, clip]{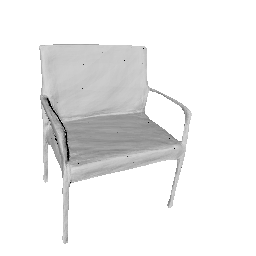} \\
        \includegraphics[width=5.5cm, trim = 0 0 0 0, clip]{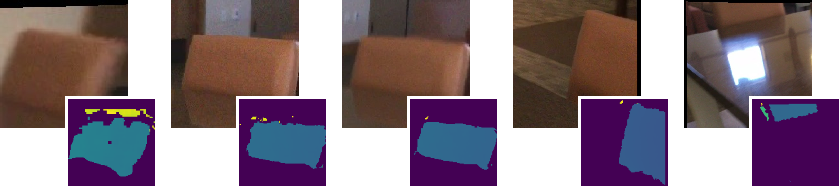} &
        \includegraphics[width=1.3cm, trim = 0 0 0 0, clip]{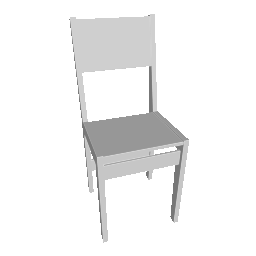} &
        \includegraphics[width=1.3cm, trim = 0 0 0 0, clip]{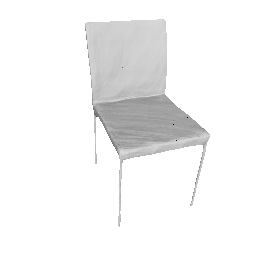} \\
        \includegraphics[width=5.5cm, trim = 0 0 0 0, clip]{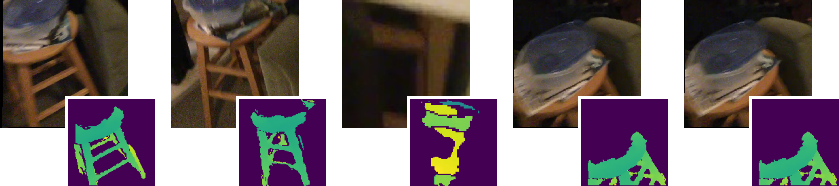} &
        \includegraphics[width=1.3cm, trim = 0 0 0 0, clip]{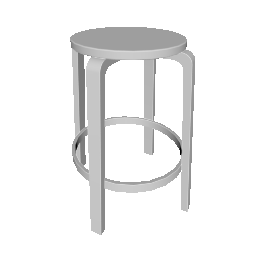} &
        \includegraphics[width=1.3cm, trim = 0 0 0 0, clip]{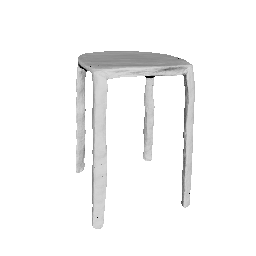} \\
        \includegraphics[width=5.5cm, trim = 0 0 0 0, clip]{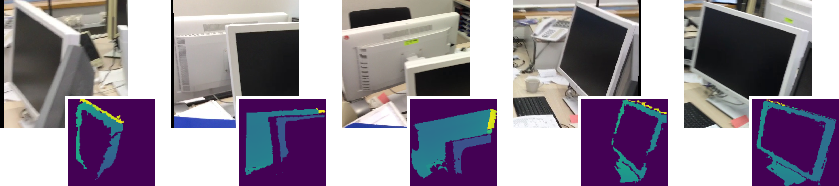} &
        \includegraphics[width=1.3cm, trim = 0 0 0 0, clip]{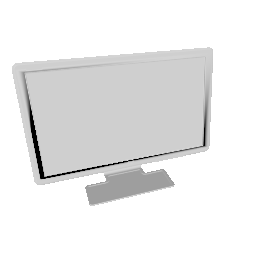} &
        \includegraphics[width=1.3cm, trim = 0 0 0 0, clip]{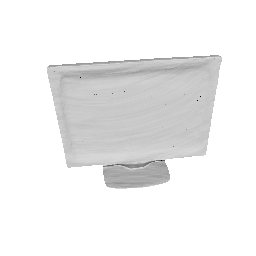} \\
        \includegraphics[width=5.5cm, trim = 0 0 0 0, clip]{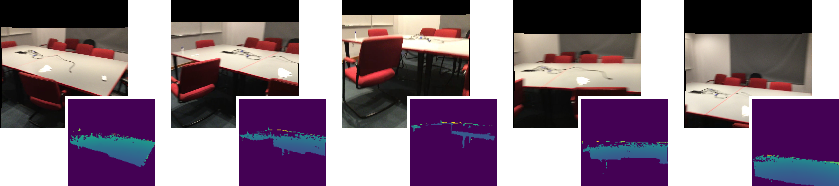} &
        \includegraphics[width=1.3cm, trim = 0 0 0 0, clip]{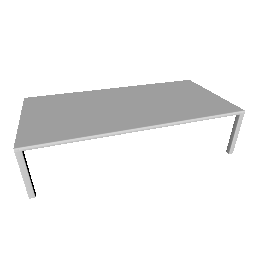} &
        \includegraphics[width=1.3cm, trim = 0 0 0 0, clip]{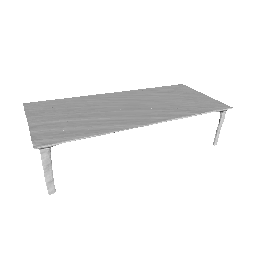} \\
    \end{tabular}
    \vspace{-0.5\baselineskip}
    \caption{Example shape and pose estimation results for ScanNet~\cite{dai2017scannet}. TransPoser can handle heavy occlusions.}
    \label{fig:scan2cad_qualitative}
    \vspace{-12pt}
\end{figure}

\Cref{tab:scan2cad_pose,tab:scan2cad_shape,fig:scan2cad_qualitative} quantitatively compares the proposed TransPoser with state-of-the-art methods. TransPoser clearly outperforms other methods including ELLIPSDF~\cite{shan2021ellipsdf} that also estimates the pose and shape code. TransPoser required 0.06 sec for each optimization step, and 0.61 sec for a total of 10 iterations with GeForce RTX 2080 Ti. This is approximately 10 times faster than those of ELLIPSDF which required 0.75 sec for its first SGD step.
Note that the exact instances used in the ELLIPSDF are not available~\cite{ellipsdf_author} and we used the test split of Scan2CAD instead. Examples in \cref{fig:scan2cad_qualitative} demonstrate that TransPoser can reconstruct plausible 3D shapes even from partial observations. These results show that our neural optimization model successfully learns to leverage the structured bias in category-level shape pose estimation.

\section{Conclusions}
In this paper, we introduced TransPoser, a novel neural optimization method for joint estimation of object shape and pose, particularly for sequentially acquired depth images. The method leverages DeepDDF which is a novel category-level neural 3D shape representation that can directly output a depth image of the object instance with consistent multi-view geometry. We showed that DeepDDF alone is an efficient and accurate shape representation that can significantly speed up the joint estimation. TransPoser, through its learned weighted optimization as forward inference and its learned momentum that extracts structured bias in category-level joint shape and pose estimation, achieves state-of-the-art accuracy as demonstrated in the extensive comparative experiments. We believe these results realize significant advances on this challenging key task for many applications in computer vision, robotics, and AR/VR.

\vspace{-12pt}
\paragraph{Limitation}
DeepDDF as a shape representation is limited to static, rigid objects and TransPoser assumes static object pose throughout its inference. We plan to expand both to handle deformable objects and dynamic relative movements of objects in our future work. The methods also are limited to depth images as inputs and do not leverage the often accompanied RGB appearance information. We plan to incorporate appearance features in the two models, which we believe will further improve the accuracy of shape reconstruction.

\vspace{-12pt}
\paragraph*{Acknowledgement}
This work was in part supported by
JSPS %KAKENHI
20H05951, % 学変
21H04893, % 基板A
JST JPMJCR20G7, % 日独仏
and RIKEN GRP.

\appendix

\renewcommand\thefigure{\thesection.\arabic{figure}}
\renewcommand\thetable{\thesection.\arabic{table}}

\section{Details of DeepDDF}
This section describes the details of DeepDDF and its experimental validation.

\subsection{Details of Network Architecture}

\begin{figure}[t]
    \centering
    \includegraphics[width=\linewidth]{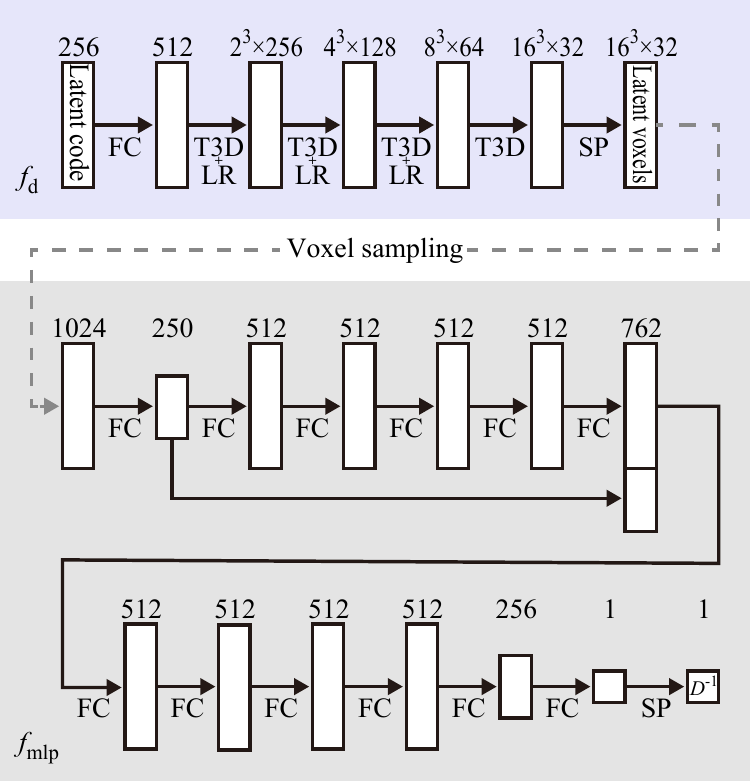}
    \caption{DeepDDF network architecture.  The rectangles denote tensors and the arrows with FC, T3D, T3D+LR, and SP labels denote a fully-connected layer, a transposed 3D convolution, a transposed 3D convolution with leaky ReLU, and a Softplus layer, respectively.}
    \label{fig:ddf_net}
\end{figure}

As shown in \cref{fig:ddf_net}, DeepDDF consists of the object instance decoder $f_\text{d}$ and the camera ray sampler $f_\text{mlp}$.
We use 256 as the dimensionality of the latent code. The decoder $f_\text{d}$ first projects the latent code to a 512 dimensional vector with a fully connected layer, and then transforms it to a $16 \times 16 \times 16 \times 32$ tensor with four 3D transposed convolutions with leaky ReLU. We set the kernel size, the stride, and the number of output channels of each 3D transposed convolution layer to 4, 2, and half the number of input channels, respectively.

The $16 \times 16 \times 16 \times 32$ tensor represents a $16 \times 16 \times 16$ latent voxel grid covering the 3D cube of edge length $1$ in which each ShapeNet CAD model is normalized to fit.
The camera ray sampler $f_\text{mlp}$ casts a viewing ray through a camera pixel of interest from the viewpoint on the unit canonical sphere towards the latent voxel grid.  It samples 32 points equally distributed along the ray at distances in the range $[1-\sqrt{3}/2, 1+\sqrt{3}/2]$ from the viewpoint.
$f_\text{mlp}$ obtains the features at the sampling points with trilinear interpolation of neighboring voxels, and concatenates the sampled features into a 1024 dimensional vector.  As shown in \cref{fig:ddf_net}, this 1024 dimensional vector is then sent to a feed-forward network consisting of 12 fully connected layers.  We concatenate the 250 dimensional vector after the first FC layer to the input of the 7th layer. The scalar output of the feed-forward network is mapped to a non-negative value by Softplus which corresponds to the inverse distance $D^{-1}$.

\subsection{Details of DeepDDF Training}
DeepDDF is trained with 200 pairs of depth and normal images synthesized from different viewpoints per instance.  Each instance is positioned so that its bounding box center coincides with the origin of the coordinate frame and its sides align with the axes.  At each epoch, for each training instance, we randomly select 1 out of the 200 pairs, and randomly sample pixels in the viewpoint to evaluate the loss function.  For efficiency, we apply importance sampling of pixels with
a probability of $0.2$ inside the object mask and with a probability of $2\times10^{-3}$ outside of it. We also dilate the object mask in advance so that pixels around the object boundary are also sampled.
We trained the networks for 3,000 epochs for the \textit{Table} category, and 10,000 epochs for the other categories. We use Adam optimizer~\cite{adam2015diederik} with a learning rate of $1.0\times10^{-4}$ and $5.0\times10^{-4}$ for the network parameters and the latent code, respectively.

\subsection{Ablation Study}

\begin{figure}[t]
    \centering
        \setlength{\tabcolsep}{0mm}
        \begin{tabular}{ccccccc}
            \footnotesize{Front} &  & \footnotesize{Back} &  & \footnotesize{Front} &  & \footnotesize{Back}\\
            \includegraphics[height=1.85cm, trim = 0 0 0 0, clip]{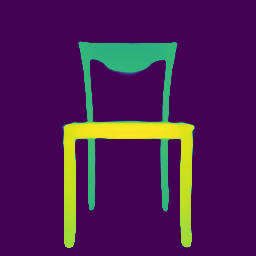} &
            \;\; &
            \includegraphics[height=1.85cm, trim = 0 0 0 0, clip]{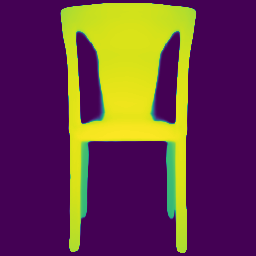} &
            \;\; &
            \includegraphics[height=1.85cm, trim = 0 0 0 0, clip]{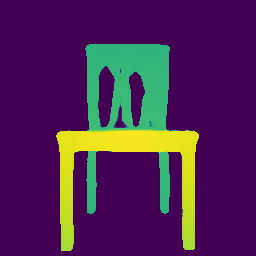} &
            \;\; &
            \includegraphics[height=1.85cm, trim = 0 0 0 0, clip]{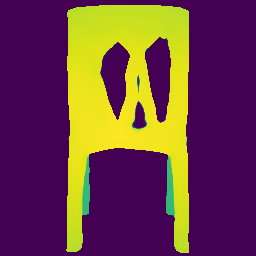} \\
            \multicolumn{3}{c}{\footnotesize{(a) w/o the latent voxel grid}} &  & \multicolumn{3}{c}{\footnotesize{(b) w/ the latent voxel grid}}
        \end{tabular}
    \vspace{-0.5\baselineskip}
    \caption{Front and back views of an object instance interpolated in the latent spaces of DeepDDF with and without the latent voxel grid.  Both shapes are interpolated as the midpoint of the latent codes of the same set of object instances in the training set. The results show that the latent voxel grid is essential to ensure multi-view geometric consistency in DeepDDF.}
    \label{fig:ddf_voxel}
\end{figure}

\Cref{fig:ddf_voxel} compares depth images reconstructed by DeepDDF and also by DeepDDF with the latent voxel grid ablated. The two networks are trained with 35 object instances (\ie, CAD models) in the \textit{Chair} category.
As we used a smaller number of object instances for this ablation study, the latent space is sparsely populated, which makes their interpolations susceptible to geometric inconsistencies.
\Cref{fig:ddf_voxel} shows depth images of an object instance interpolated from the same pair of training samples using DeepDDF and DeepDDF without the latent voxel grid. By comparing the front and the back views, we can observe that DeepDDF without the latent voxel grid generates inconsistent depth-maps, while the full DeepDDF generates plausible object geometry.  These results clearly show that the latent voxel grid is essential for generating objects shape with multi-view geometric consistency.

\subsection{Details of Shape Reconstruction Experiment}

\label{sec:non_liner_opt}
In the shape reconstruction experiment reported in Table 2 and Figure 5 of the main text, we reconstruct the shape from a single-view depth image given known camera and object poses.
We optimize the latent code $\vct{z}$ to minimize the differences between the observed and the synthesized inverse distance images $\vct{I}_{\rm obs}$, $\vct{I}_{\rm est}(\vct{z})$
\begin{equation}
 \vct{z}^{*} = \argmin_{\vct{z}} \lambda_{\rm{d}} |\vct{I}_{\rm obs}-\vct{I}_{\rm est}(\vct{z})| + \lambda_{\rm{r}} ||\vct{z}||\,,
\label{opt}
\end{equation}
where $\lambda_{\rm{d}}$ and $\lambda_{\rm{r}}$ control the weighting between the prediction error and the regularization on the latent code.
$\lambda_{\rm{d}}$ and $\lambda_{\rm{r}}$ are set to $10.0$ and $1.0$, respectively.
We use Adam Optimizer with a learning rate $1.0 \times 10^{-3}$, and perform 200 steps of optimization.
For DIST~\cite{liu2020dist}, we used the pre-trained models provided by the original authors, and set the hyperparameter values to those used in the original implementation except for the following changes. We omitted the normal error in the shape optimization according to \cref{opt}.
We also set the number of sampling points $K$ used for rendering depth images by DIST to 1, as we empirically found the original setting, namely $K=3$, noticably decreased accuracy.

For the \textit{Chair} and \textit{Table} categories, we use the test split of DeepDDF which has 832 and 1,216 CAD models respectively. We render a $256 \times 256\text{px}$ depth image of $60^\circ$ field-of-view from a random viewpoint on the unit canonical sphere for each instance.
To compute the reconstruction accuracy, we generated a point cloud corresponding to the estimated latent code by integrating DeepDDF depth images rendered from 6 viewpoints distributed uniformly around the object.

\section{Details of TransPoser}
This section describes the details of TransPoser and its experimental validation on synthetic and real-world data.

\subsection{Details of Parameters and Loss Functions}
We represent the translation and scale by 3-dimensional vectors $\vct{p}, \vct{s} \in \mathbb{R}^3$, respectively.
 The 3D rotation is represented by two 3-dimensional vectors $\vct{d}_\text{g}, \vct{d}_\text{r} \in \mathbb{R}^3$ that represents the two axes of the normalized object coordinate system as done in FS-Net~\cite{Chen_2021_CVPR}.
The shape is represented by the latent code $\vct{z} \in \mathbb{R}^{256}$ of DeepDDF.

 We update the current estimate $\vct{p}^t, \vct{d}_\text{g}^t, \vct{d}_\text{r}^t, \vct{s}^t$, and $\vct{z}^t$ with their updates $\Delta \vct{p}^t, \Delta \vct{d}_\text{g}^t, \Delta \vct{d}_\text{r}^t, \Delta \vct{s}^t$, and $\Delta \vct{z}^t$, respectively (\ie, the outputs of TransPoser),
\begin{align}
\vct{p}^{t+1} &= \vct{p}^t + \Delta \vct{p}^t\,, \\
\vct{d}_\text{g}^{t+1} &= \text{normalize}(\vct{d}_\text{g}^t + \Delta \vct{d}_\text{g}^t)\,, \\
\vct{d}_\text{r}^{t+1} &= \text{normalize}(\vct{d}_\text{r}^t + \Delta \vct{d}_\text{r}^t)\,, \\
\vct{s}^{t+1} &= \vct{s}^t \circ \Delta \vct{s}^t\,, \\
\vct{z}^{t+1} &= \vct{z}^t + \Delta \vct{z}^t\,,
\end{align}
where $\text{normalize}(\cdot)$ normalizes the given vector to unit magnitude, and $\circ$ denotes the Hadamard product.

The loss of TransPoser is the discrepancy between the updated estimates and their corresponding ground truth values.
We calculate the difference between each estimate $\vct{p}^{t+1}, \vct{d}_\text{g}^{t+1}, \vct{d}_\text{r}^{t+1} \vct{s}^{t+1}$, and $\vct{z}^{t+1}$ and their ground truth $\vct{p}_{\text{gt}}, \vct{d}_{\text{g}_{\text{gt}}}, \vct{d}_{\text{r}_{\text{gt}}} \vct{s}_{\text{gt}}$, and $\vct{z}_{\text{gt}}$, respectively,
\begin{align}
\mathcal L_{p} &= \text{MSE}(\vct{p}^{t+1}, \vct{p}_{\text{gt}})\,, \\
\mathcal L_{g} &= 1 - \text{COSSIM}(\vct{d}_\text{g}^{t+1}, \vct{d}_{\text{g}_{\text{gt}}})\,, \\
\mathcal L_{r} &= 1 - \text{COSSIM}(\vct{d}_\text{r}^{t+1}, \vct{d}_{\text{r}_{\text{gt}}})\,, \\
\mathcal L_{s} &= \text{MSE}(\vct{s}^{t+1}, \vct{s}_{\text{gt}})\,, \\
\mathcal L_{z} &= \text{MSE}(\vct{z}^{t+1}, \vct{z}_{\text{gt}})\,,
\end{align}
where $\text{MSE}$ and $\text{COSSIM}$ denote the mean squared error and the cosine similarity, respectively.
We train TransPoser with the sum of the errors over all optimization steps
\begin{equation}
\mathcal L = \sum_{t=1}^{T} (\lambda_p \mathcal L_{p} + \lambda_g \mathcal L_{g}+\lambda_r \mathcal L_{r}+\lambda_s \mathcal L_{s} + \lambda_z \mathcal L_{z})\,, \\
\end{equation}
where $\lambda_p, \lambda_g, \lambda_r, \lambda_s, \lambda_z$ control the weighting between each estimation, and $T$ denotes the total number of iterations.

\subsection{Momentum with Cross-attention}
TransPoser effectively learns momentum of the iterative joint shape and pose estimation in the sense that it incorporates an adaptive amount of past estimates and combines it with the current estimate to output an adjusted update. This is achieved with the Transformer decoder of TransPoser, which calculates updates based on the prediction errors at the current time step and also the previous time step. Table 3 of the main text shows that the decoder improves the reconstruction accuracy. We believe this learned momentum with cross-attention can be essential for neural optimization and plan to explore its use in other optimization tasks.

\subsection{Details of Network Architecture}
TransPoser consists of a ResNet-18 backbone, three-layer Transformer encoder and decoder, and five two-layer MLP heads.
We average pool the output of the 5th convolution block of ResNet18 to a 512 dimensional vector.
We concatenate the three vectors corresponding to the observation, the prediction, and their differences, and project them to 256 dimensions with a fully connected layer to input to the Transformer.
For this, we use three independent layers to obtain the encoder tokens $\{\vct{e}^{\ast t}_k\}_{k=1}^{K^t}$, the decoder tokens $\{\vct{m}^{t-1}_k\}_{k=1}^{K^{t-1}}$, and the viewing condition encoding of the current view $\vct{m}^t_{K^t}$.
We embed the 7D camera pose to a token with a learnable vector $\vct{W}_{\text{PE}} \in \mathbb{R}^{7 \times 256}$.

TransPoser has three layers of attention in both the encoder and the decoder.
We set the dimension of the outputs and inputs to $d_{\text{model}}=256$, the dimension of the feed-forward network model to $d_{\text{ff}}=1024$, the number of the multi-head attentions as $h=8$, and probability of dropout to $P_\text{drop}=0.1$.

The five MLP heads output the updates $\Delta \vct{p}^t$, $\Delta \vct{d}_\text{g}^t$, $\Delta \vct{d}_\text{r}^t$, $\Delta \vct{s}^t$, and $\Delta \vct{z}^t$, respectively.
We set the dimension of each layer to 256.
We use ReLU and Leaky ReLU as the activation functions of the encoder and decoder, and the MLP heads, respectively.
We adopt Softplus for the final output of the MLP head for the scale $\Delta \vct{s}^t$.

\subsection{Details of Experiments with Synthetic Data}
This section describes the details of producing the synthetic data and training using it.
\begin{figure}[t]
    \centering
        \setlength{\tabcolsep}{0mm}
        \begin{tabular}{cc}
            \includegraphics[height=3.cm, trim = 65 70 35 40, clip]{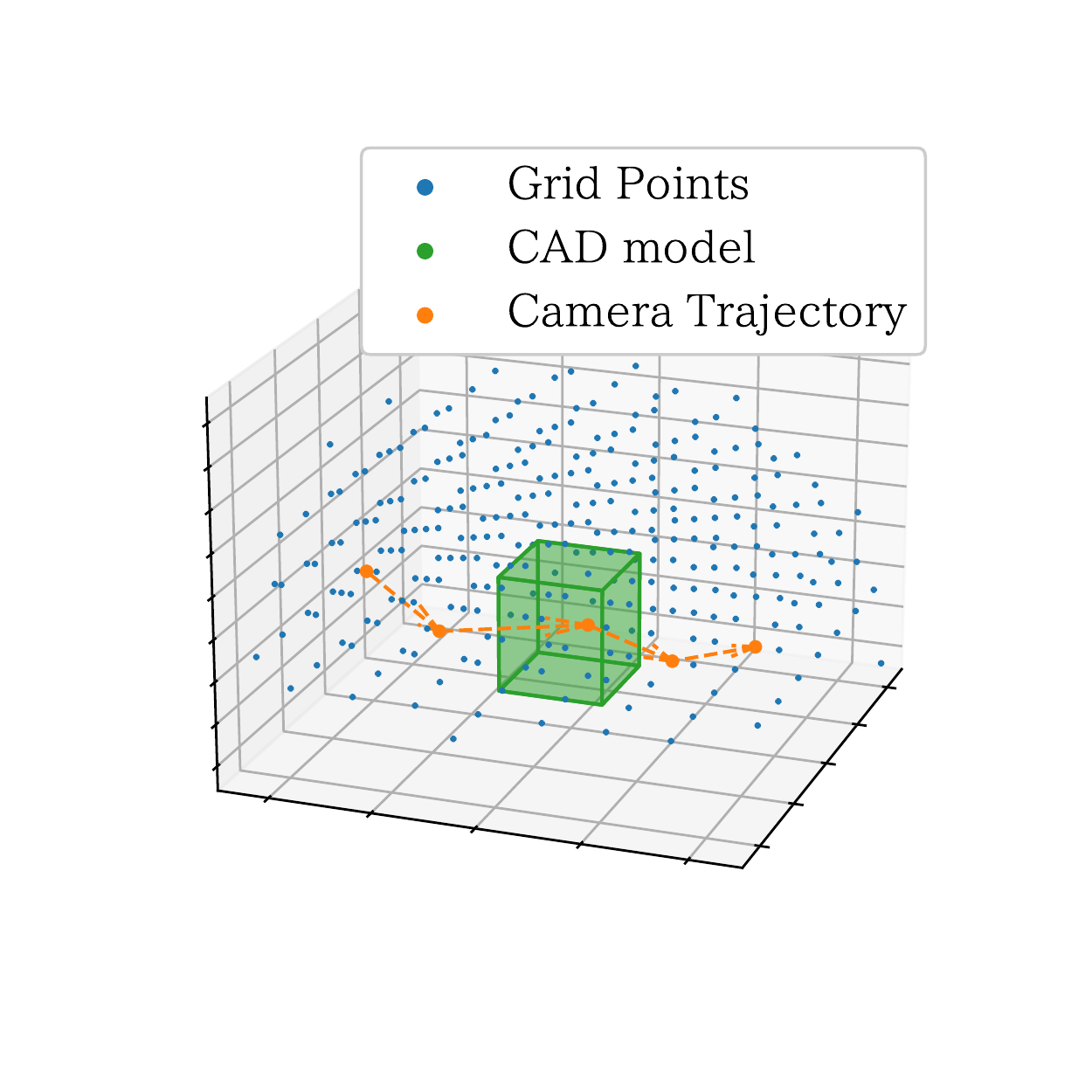} &
            \includegraphics[height=2.7cm, trim = 0 0 0 0, clip]{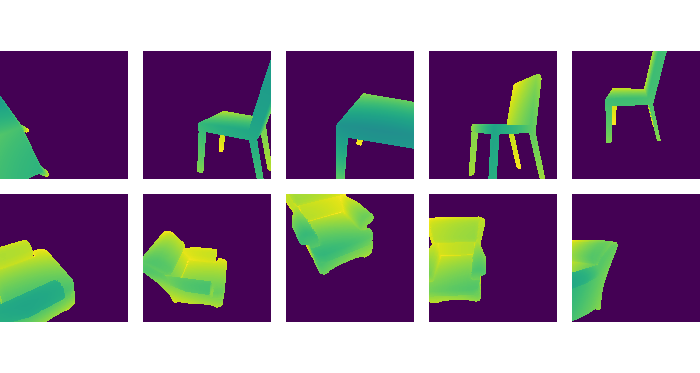} \\
            \footnotesize{(a) Generated Viewpoints}\;\;\; & (b) \footnotesize{Sample Rendered Images}\\
             & \vspace{-18pt}\\
        \end{tabular}
    \caption{Examples of the randomly generated camera trajectories (a) and the sets of five synthetic observations (b).}
    \vspace{-0.5\baselineskip}
    \label{fig:syndata_samples}
\end{figure}
\subsubsection{Data Preparation}
As shown in \Cref{fig:syndata_samples}, we generate camera trajectories by randomly sampling points on a regular grid inside the upper hemisphere around the object.
The center of the sphere coincides with that of the object instance's bounding box, and its radius is 1.5.
The interval of the grids is set to 0.3.
The viewpoint starts at a random grid point, and randomly moves up to 2 grid points for each movement.
We randomize the view direction while ensuring that at least $2\%$ of the pixels capture the target object.
The scale is randomized by a factor within the range of $[0.8, 1.2]$.

Supervised training of TransPoser requires ground truth latent code of DeepDDF for each instance.  As those latent codes are available only for the training set of DeepDDF, we split it into 100, 512, 100, and 1,536 training instances and 100, 785, 100, and 1,216 validation instances to train TransPoser for the \textit{Cabinet}, \textit{Chair}, \textit{Display}, and \textit{Table} categories, respectively. We used the test set of DeepDDF as the test set for TransPoser, too.

For each epoch of training, we prepare a trajectory consisting of a set of five images for each object instance, and train the network to optimize the pose and the shape for 10 iterations. For validation and testing, we prepare 8 and 16 trajectories for each instance, respectively.

\begin{figure}[t]
    \centering
    \setlength{\tabcolsep}{0mm}
    \begin{tabular}{ccc}
        \includegraphics[height=3.67cm, trim = 8  0 12 0, clip]{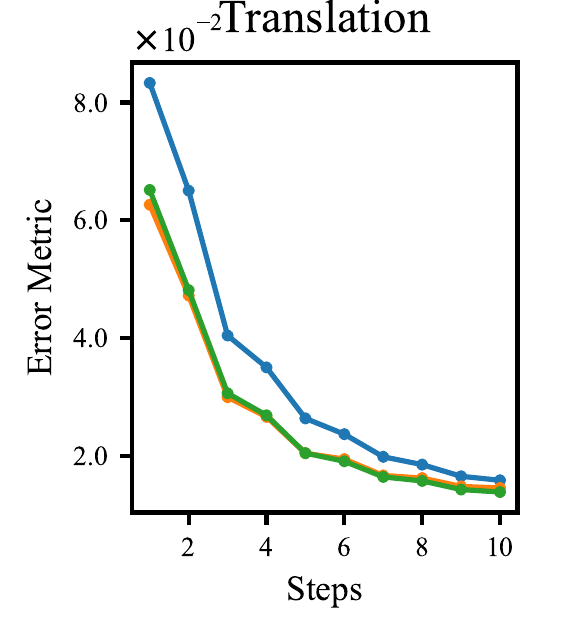} &
        \includegraphics[height=3.67cm, trim = 20 0 12 0, clip]{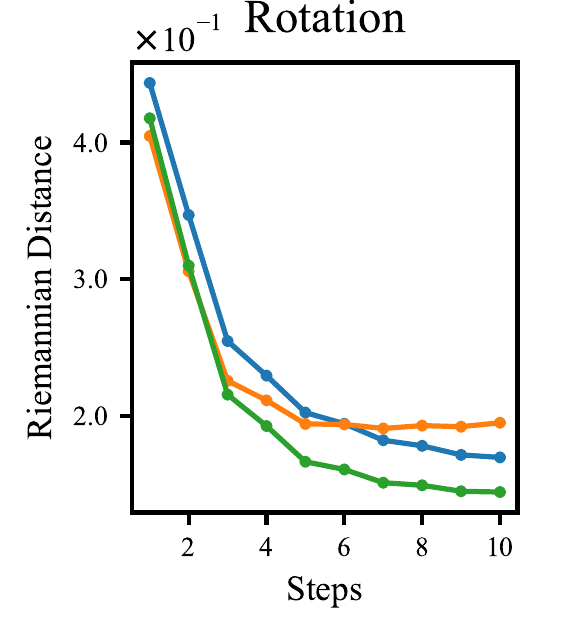} &
        \includegraphics[height=3.67cm, trim = 20 0 12 0, clip]{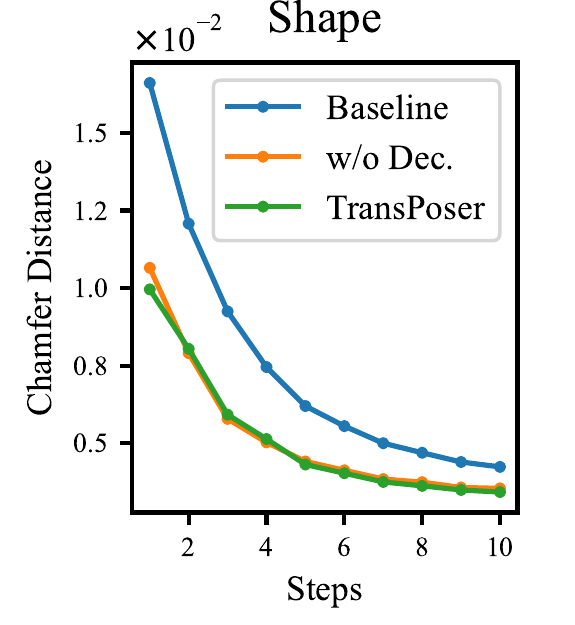} \\
    \end{tabular}
    \vspace{-1.2\baselineskip}
    \caption{Comparison of convergence speeds. The plots show, from left to right, the estimation errors of translation, rotation, and shape at each optimization step, respectively. The error metrics are those of \Cref{tab:sufcace_map}.  The results show that the proposed TransPoser achieves fastest and most accurate reconstruction.}
    \label{fig:dfnet_speed}
\end{figure}

\begin{figure}[t]
    \centering
    \setlength{\tabcolsep}{0mm}
    \begin{tabular}{cccc}
        \vspace{-13pt} & & \\
        \footnotesize{\thead{Observations\\ and GT}} & \footnotesize{\thead{Baseline\\ based on \cite{ma2020deep} }} & \footnotesize{w/o Dec.} & \footnotesize{TransPoser}\\
        \includegraphics[height=1.6cm, trim = 0 0 0 0, clip]{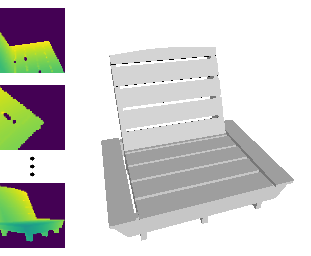} &
        \includegraphics[height=1.6cm, trim = 0 0 0 0, clip]{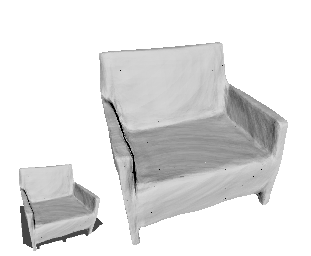} &
        \includegraphics[height=1.6cm, trim = 0 0 0 0, clip]{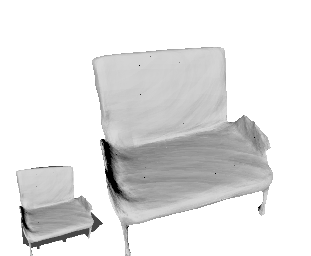} &
        \includegraphics[height=1.6cm, trim = 0 0 0 0, clip]{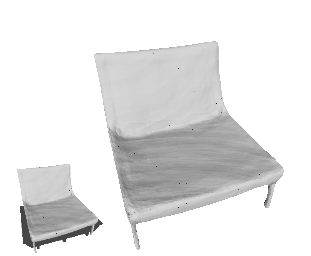} \\
        \includegraphics[height=1.6cm, trim = 0 0 0 0, clip]{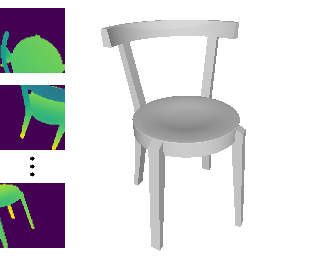} &
        \includegraphics[height=1.6cm, trim = 0 0 0 0, clip]{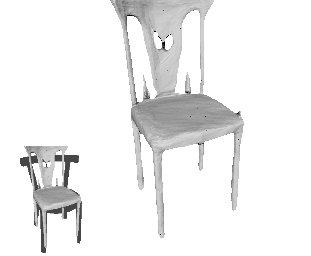} &
        \includegraphics[height=1.6cm, trim = 0 0 0 0, clip]{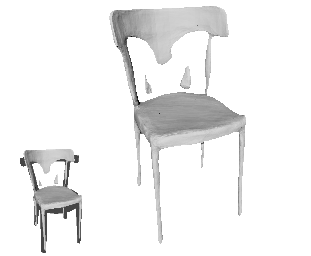} &
        \includegraphics[height=1.6cm, trim = 0 0 0 0, clip]{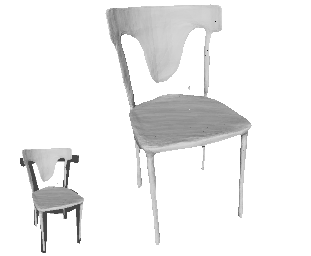} \\
        \includegraphics[height=1.6cm, trim = 0 0 0 0, clip]{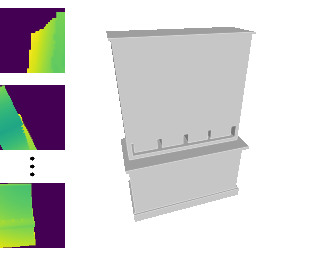} &
        \includegraphics[height=1.6cm, trim = 0 0 0 0, clip]{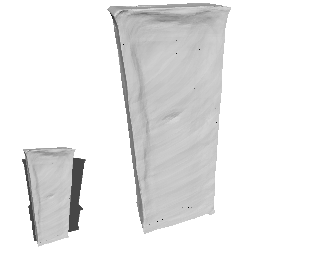} &
        \includegraphics[height=1.6cm, trim = 0 0 0 0, clip]{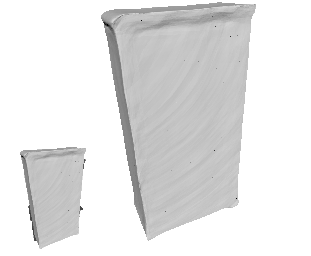} &
        \includegraphics[height=1.6cm, trim = 0 0 0 0, clip]{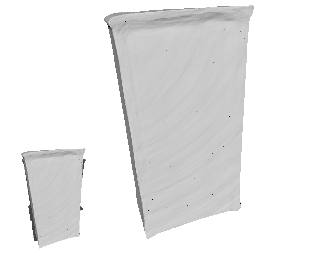} \\
    \end{tabular}
    \vspace{-0.5\baselineskip}
    \caption{Examples of reconstructed shapes by a baseline method, TransPoser without the decoder, and the full TransPoser. The inset images show the estimated shape (gray) over the ground truth silhouette (black) to visualize the difference.  The results demonstrate the importance of self-attention which weighs the different views.}
    \label{fig:dfnet_qlt}
\end{figure}

\subsubsection{Training Details}
We initialize the translation with uniform random noise of $\mathcal{U}[-0.3, 0.3]$, the scale by $\mathcal{U}[0.7, 1.3]$, and the rotation by a random orientation.
The shape is initialized with the mean latent code of the training samples.

We trained TransPoser for 512 epochs using Adam optimizer with a learning rate of $1.0\times10^{-4}$, and use the one with the minimum validation loss. In the first 80 epochs, we trained TransPoser to optimize the the pose and shape not in 10 iterations but in 5 iterations, as we experimentally found that this stabilizes the training process.  We used $\lambda_p=10.0, \lambda_g=1.0, \lambda_r=1.0, \lambda_s=1.0, \lambda_z=1.0$ for the \textit{Cabinet}, \textit{Chair}, and \textit{Display} categories. For the \textit{Table} category, we used $\lambda_r=0.1$.  We evaluated the rotation loss by considering the symmetric shape ambiguity~\cite{Chen_2021_CVPR}.
\begin{table}[t]
    \centering
    \begingroup
    \scriptsize
    \setlength{\tabcolsep}{1.43mm}
    \begin{tabular}{l||ccc|cccccc}
        \toprule[1.5pt]
        \multirow{2}{*}{Model} &
        \multicolumn{3}{c|}{Class avg.} & \multicolumn{3}{c}{\textit{Cabinet}} & \multicolumn{3}{c}{\textit{Chair}} \\
         & T & R & S & T & R & S & T & R & S  \\
        \hline
        w/o NOCS & 1.58 & 1.18 & 3.52 & 1.87 & 1.69 & 4.09 & 1.30 & 0.66 & 2.95 \\
        w/o Camera PE & 1.29 & \textbf{0.98} & 3.12 & 1.40 & 1.38 & 3.62 & 1.19 & \textbf{0.58} & 2.61 \\
        w/o $\vct{m}^t_{K^t}$ & 1.35 & 1.01 & 3.13 & 1.59 & \textbf{1.35} & 3.52 & 1.10 & 0.67 & 2.74 \\
        \rowcolor[rgb]{0.93,1.0,0.87} \textbf{TransPoser} & \textbf{1.23} & 1.00 & \textbf{3.03} & \textbf{1.35} & 1.39 & \textbf{3.47} & \textbf{1.10} & 0.61 & \textbf{2.58} \\
        \bottomrule[1.5pt]
    \end{tabular}
    \vspace{-0.5\baselineskip}
    \endgroup
    \caption{Ablation study of the NOCS representation, the positional encoding of the camera poses, and the addition of $\vct{m}^t_{K^t}$ to the input of the decoder. Metrics for translation (T), rotation (R), and shape (S) are the mean squared error $(10^{-2})$, Riemannian distance $(10^{-1})$, and Chamfer distance $(10^{-3})$, respectively. The results show that each component are independently essential to TransPoser.}
    \label{tab:sufcace_map}
\end{table}

\subsection{Details of Experiments with Real Data}
This section describes the details of prepossessing and fine-tuning for real data.
\subsubsection{Data Preparation}
We followed the training, validation, and test splits defined in Scan2CAD, and used the CAD models aligned with the objects in the real images.
The latent codes of DeepDDF required for supervised training of TransPoser were computed by fitting DeepDDF nonlinearly to the CAD models.
We manually removed several objects with incorrect instance masks from the training and validation splits.
For a fair performance assessment, we also removed such objects from the test set when evaluating the shape reconstruction accuracy.

We initialized the translation with the median of the 3D point clouds constructed from the ones inside the instance mask of the first frame depth image. The rotation is initialized with a random direction perpendicular to the floor. The scale is set so that the diagonal of the object bounding box is $1 \mathrm{m}$ for the \textit{Chair} and \textit{Display} categories, and $2 \mathrm{m}$ for the \textit{Cabinet} and \textit{Table} categories. The initial shape is the mean of the latent codes of all the DeepDDF training samples in the category.

\subsubsection{Training Details}
We fine-tuned TransPoser pre-trained with the synthetic data, and tested the model that achieved the highest validation accuracy.  We evaluate the accuracy with the pose evaluation metrics proposed by Scan2CAD.
We use Adam optimizer with a learning rate of $5.0\times10^{-6}$ for the \textit{Cabinet} and \textit{Display} categories, and $1.0\times10^{-5}$ for the \textit{Chair} and \textit{Table} categories.
We used the same weights for the losses as in the the synthetic data experiment, and evaluated the rotation loss by considering the symmetric shape ambiguity~\cite{Chen_2021_CVPR}.

\begin{figure}[t]
    \centering
    \setlength{\tabcolsep}{0mm}
    \begin{tabular}{ccc}
        \vspace{-13pt} & & \\
        \footnotesize{Observations} & \footnotesize{GT} & \footnotesize{Est.} \\
        \includegraphics[height=1.15cm, trim = 0 -25 0 0, clip]{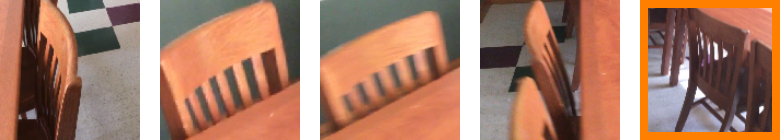} &
        \includegraphics[height=1.3cm, trim = 0 0 0 0, clip]{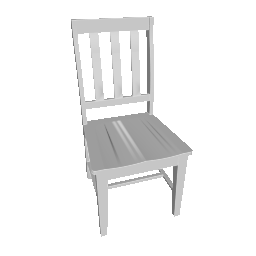} &
        \includegraphics[height=1.3cm, trim = 0 0 0 0, clip]{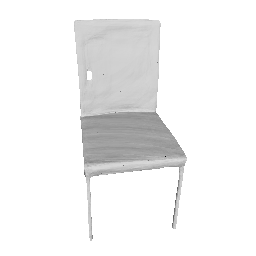} \\
        \includegraphics[height=1.15cm, trim = 0 -25 0 0, clip]{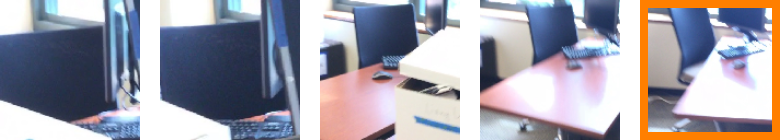} &
        \includegraphics[height=1.3cm, trim = 0 0 0 0, clip]{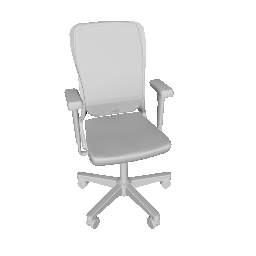} &
        \includegraphics[height=1.3cm, trim = 0 0 0 0, clip]{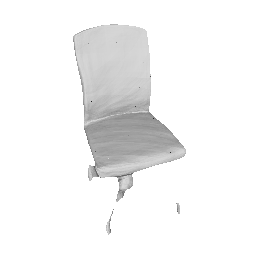} \\
        \includegraphics[height=1.15cm, trim = 0 -25 0 0, clip]{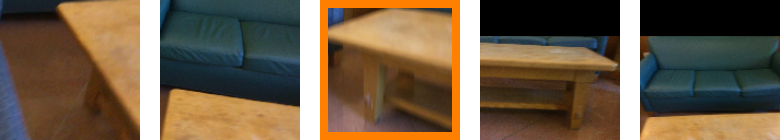} &
        \includegraphics[height=1.3cm, trim = 0 0 0 0, clip]{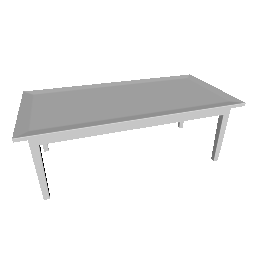} &
        \includegraphics[height=1.3cm, trim = 0 0 0 0, clip]{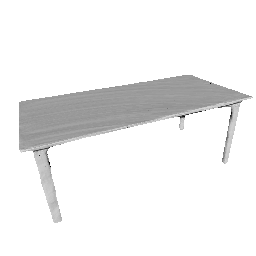} \\
    \end{tabular}
    \vspace{-0.5\baselineskip}
    \caption{Visualization of the most attended views (images with the orange borders). They show that TransPoser effectively attends to informative views with rich silhouette information.}
    \label{fig:attn}
\end{figure}

\subsection{Additional Ablation Studies}

\paragraph{Convergence speed}
\Cref{fig:dfnet_speed,fig:dfnet_qlt} show results of comparing Transposer with a baseline method based on \cite{ma2020deep} and TransPoser without the decoder for convergence speed and qualitative results, respectively. \Cref{fig:dfnet_speed} demonstrates quantitatively that the proposed TransPoser achieves the highest accuracy in most of the metrics while achieving a faster convergence. TransPoser without the decoder also converges as fast as TransPoser, but is less accurate as evident in the rotation error.
As shown in \Cref{fig:dfnet_qlt}, the baseline model fails to reconstruct the object shape in the observations.
This is mainly because the baseline methods cannot coherently integrate partial observations from each view.

We also evaluate the contributions of the NOCS representation~\cite{wang2019normalized}, the positional encoding of the camera poses, and adding $\vct{m}^t_{K^t}$ to the input of the decoder when $K^{t} = K^{t-1} + 1$. We evaluated with the \textit{Chair} and \textit{Cabinet} categories, and set the number of total optimization steps to 5.

\Cref{tab:sufcace_map} shows the results of TransPoser without NOCS representation, TransPoser without the positional encoding of the camera poses, and TransPoser without $\vct{m}^t_{K^t}$ in the decoder, in comparison with the full TransPoser. These results clearly show that each of the components independently contributes to the shape and pose estimation accuracy.

\Cref{fig:attn} visualizes, with the orange borders, the views that TransPoser pays the most attention to among the five input observations.
The visualization is based on the attention of the last cross-attention layer before computing the update.
The results show qualitatively that TransPoser properly focuses on the viewpoints that are informative for estimating the pose and shape, such as the images captured from oblique angles which contain rich silhouette information.

\balance
{\small
\bibliographystyle{ieee_fullname}
\bibliography{egbib}
}

\end{document}